\documentclass[runningheads]{llncs}

\PassOptionsToPackage{table}{xcolor}

\usepackage{eccv}

\usepackage{eccvabbrv}

\usepackage{tabularray}
\usepackage{pifont}
\newcommand{\circled}[1]{\ding{\numexpr 181 + #1\relax}}
\usepackage[accsupp]{axessibility}  
\usepackage{graphicx}
\usepackage{booktabs}
\usepackage[most]{tcolorbox}
           
\usepackage{placeins} 
\usepackage{multirow}
\usepackage{makecell} 
\usepackage{lipsum}
\usepackage{soul}
\sethlcolor{yellow}
\usepackage[normalem]{ulem}
\usepackage{xcolor}    
\usepackage{amssymb}   
\usepackage{pifont}    
\usepackage{wrapfig} 
\usepackage{array}      
\usepackage{siunitx}    
\usepackage{adjustbox}
\usepackage{xcolor}
\usepackage{colortbl}

\usepackage{listings}
\lstdefinestyle{jsonstyle}{
    basicstyle=\ttfamily\footnotesize,
    breaklines=true,
    frame=single,
    backgroundcolor=\color{white},
    keywordstyle=\color{blue},
    stringstyle=\color{teal}
}

\usepackage{hyperref}

\usepackage{orcidlink}

\newcommand{\correct}{\textcolor{Green}{\checkmark}} 
\newcommand{\wrong}{\textcolor{Red}{\ding{55}}}      
\newcommand{\ours}{\texttt{\textbf{ForeSea}}}
\newcommand{\oursbench}{\texttt{\textbf{ForeSeaQA}}}
\newcommand{\equalfirst}{\textsuperscript{*}}
\newcommand{\equalsecond}{\textsuperscript{\textdagger}}
\newcommand{\qualnote}{\textsuperscript{\textdaggerdbl}}

\definecolor{lightgreen}{RGB}{225,245,230}
\definecolor{lightyellow}{RGB}{255,227,74}  
\definecolor{lightred}{RGB}{255,235,220}  

\begin{document}
\title{\ours{}: AI Forensic Search with Multi-modal Queries for Video Surveillance}

\authorrunning{H.Park and Y.Li.~et al.}
\titlerunning{ForeSea}

\author{
Hyojin Park\inst{1}\equalfirst \and
Yi Li\inst{1}\equalfirst \and \\
Janghoon Cho\inst{1}\equalsecond \and
Sungha Choi\inst{2}\equalsecond\qualnote \and
Jungsoo Lee\inst{1}\equalsecond \and \\
Taotao Jing\inst{1} \and
Shuai Zhang\inst{1} \and
Munawar Hayat\inst{1} \and 
Dashan Gao\inst{1} \and \\
Ning Bi\inst{1} \and
Fatih Porikli\inst{1}
}

\institute{Qualcomm AI Research, San Diego, CA, USA \and
Kyunghee University, Gyeonggi, South Korea\\ }

\maketitle

\begin{abstract}
Despite decades of work, surveillance still struggles in searching and reasoning about specific targets across long, multi-camera videos. 
Existing methods---tracking, retrieval, and video LLMs---require heavy manual filtering, capture only shallow attributes, and fail at temporal understanding. 
Prior benchmarks are also limited to basic retrieval and question answering, without addressing real-world challenges that often involve multimodal queries and temporal grounding 
(e.g., “When did this person join the fight?” with the person’s image). 
To address this gap, we introduce \oursbench{}, a new benchmark specifically designed for video QA with image‑and‑text queries and timestamped annotations of key events. 
The dataset consists of long‑horizon surveillance footage paired with diverse multimodal questions, enabling systematic evaluation of retrieval, temporal grounding, and multimodal reasoning in realistic forensic conditions.
Not limited to this benchmark, we propose \ours{}, an AI forensic search system with a 3‑stage, plug‑and‑play pipeline. (1) A tracking module filters irrelevant footage; (2) a multimodal embedding module indexes the remaining clips; and (3) during inference, the system retrieves top‑K candidate clips for a video LLM to answer queries and localize events. On \oursbench{} benchmark, \ours{} improves accuracy by 3.1 points and temporal IoU by 10.1 points over prior retrieval-augmented baselines.
To our knowledge, \oursbench{} is the first benchmark to support complex multimodal queries with precise temporal grounding, and \ours{} is the first VideoRAG system built to excel in this setting.

\end{abstract}    
\section{Introduction}
\label{sec:intro}
{
\let\thefootnote\relax\footnotetext{{
\vspace{-0.2em}
\noindent \equalfirst\ Equal contribution as first authors.\\
\vspace{-0.2em}
\noindent \equalsecond\ Equal contribution as second authors.\\
\noindent \qualnote\ This work was done while the author was at Qualcomm.
}}
}

\begin{figure}[t]
  \centering
   \includegraphics[width=0.85\linewidth]{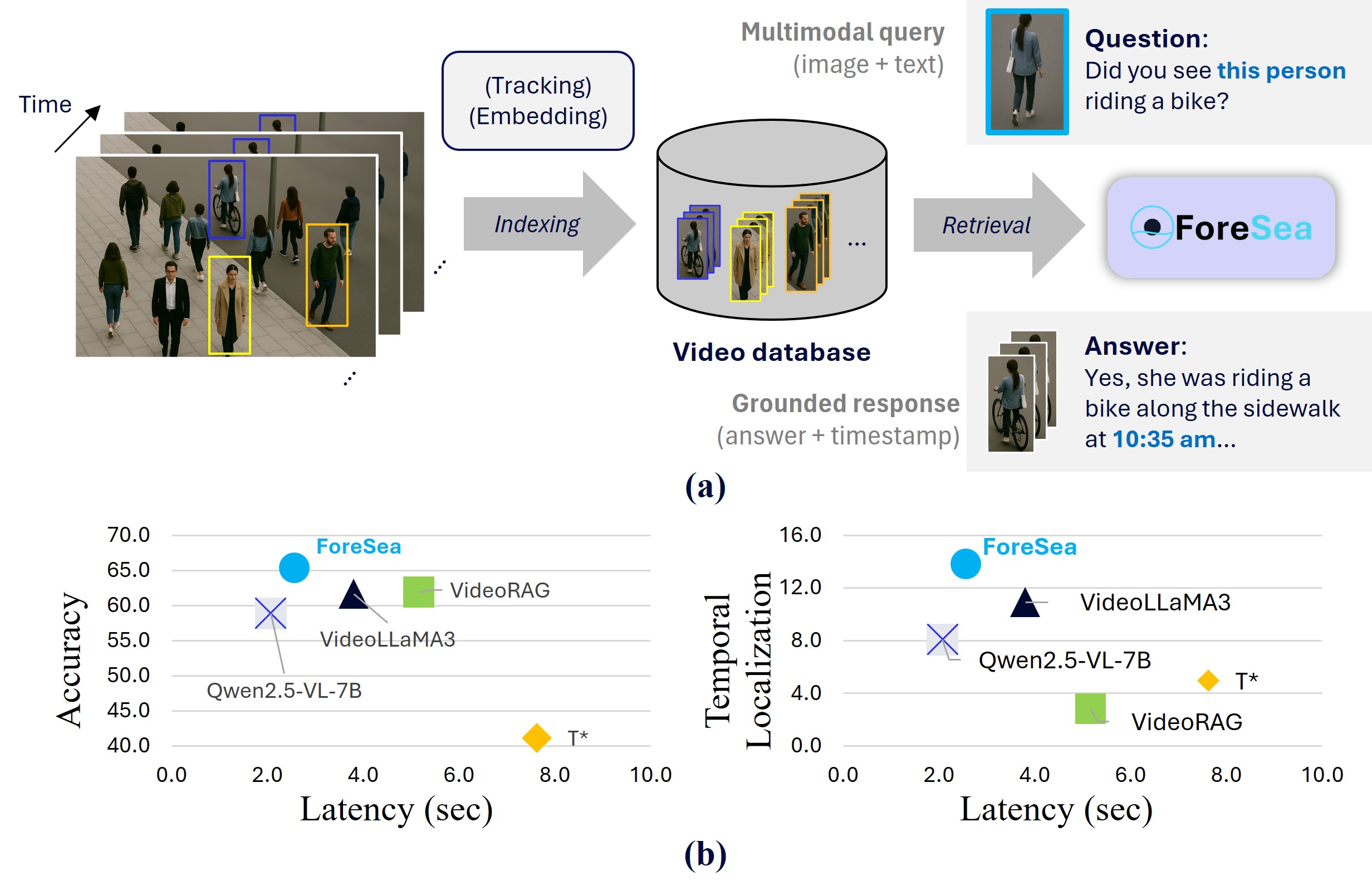}

   \caption{\textbf{AI Forensic Search with \ours{}.} Our proposed framework for long surveillance videos supports complex \emph{multimodal queries} (e.g., a reference image combined with a text question) and leverages a person-centric multimodal database to efficiently retrieve and generate \emph{temporally grounded} answers.}
   \label{fig:intro}
   \vspace{-5mm}
\end{figure}

Recent large multimodal models (LMMs) have made rapid progress in long-form video analysis, driven by advances in general video understanding~\cite{chen2023videollm,zhang2024llava,zhang2025videollama}, temporal grounding~\cite{wang2024grounded,wang2025time}, and complex reasoning~\cite{feng2025video,cheng2025video}. 
These skills are crucial for applications to video surveillance analysis~\cite{sultani2018real,yuan2023surveillance,liu2025surveillancevqa}, which requires finding specific people, objects, or events of interest across hours or even days of videos captured by multiple cameras.

Existing surveillance systems have traditionally relied on object detection and tracking pipelines~\cite{zhang2022bytetrack, pang2021qdtrack, zhou2020centertrack, bergmann2019tracktor} to process large-scale video data.
While this enables basic analytics such as counting and searching, it struggles with recognizing complex activities, detecting various anomalies, and achieving a holistic understanding of long videos.
Each of these tasks often involves substantial human effort, including reviewing retrieved footage, and gathering visual evidence to draw conclusions.

To mitigate this manual effort, recent approaches use CLIP-based models, vision-language models (VLMs) and retrieval-augmented generation (RAG) ~\cite{li2017person, luo2021clip4clip, cao2024empirical,wang2024videoagent, ren2025videorag, sagare2024videorag }. 
However, they still face three key limitations: (1) it only supports text queries, cannot handle multimodal queries; (2) the provided answers are lack of temporal evidence reasoning; and (3) VLM model's accuracy and length of input are constrained in long video.
These limitations highlight the need for practical surveillance tasks, such as answering \emph{multimodal} queries with \emph{temporally grounded} evidence (Fig.~\ref{fig:intro}). 
For example:
\begin{tcolorbox}[colback=blue!10!white, colframe=blue!10!white,left=4pt, right=4pt, top=4pt, bottom=4pt, boxrule=0pt, boxsep=1pt, arc=2pt]
\footnotesize
\textbf{Q}: ``Did you see \emph{this person} riding a bike?'' + an image of the individual

\textbf{A}: ``Yes, riding a bike \emph{at 10:35 am} along the sidewalk.'' + a trimmed video clip.
\end{tcolorbox}


While such tasks are essential for real-world scenarios, they remain absent from existing research.
To bridge this gap, we introduce \textbf{\oursbench{}}, the first benchmark for multimodal, temporally grounded video question answering in surveillance.
\oursbench{} is built from UCF-Crime~\cite{sultani2018real} videos using a semi-automated data engine that extracts person entities from dense captions~\cite{yuan2023surveillance} and visually grounds them via a multimodal LLM.
The AI then generates QA pairs with precise temporal annotations across six subtasks: \emph{search}, \emph{activity}, \emph{event}, \emph{temporal}, \emph{counting}, and \emph{anomaly}.
Crucially, person-specific questions use \emph{multimodal} queries—a reference image of the individual with the question text—to reflect real forensic workflows.
All QA pairs are manually verified for validity, unambiguity, and temporal-grounding correctness.
To our knowledge, \oursbench{} is the first benchmark to jointly evaluate multiple-choice accuracy and temporal localization under both text-only and multimodal query conditions in surveillance.


We further present \textbf{\ours{}}, a simple yet strong multimodal RAG framework combining three off-the-shelf components: (i) a person tracker that segments long videos into person-centric clips, drastically reducing the search space; (ii) a multimodal encoder that indexes clips in a unified image-text embedding space for text and image-text retrieval; and (iii) a videoLMM that reasons over the top-$K$ retrieved clips to produce a temporally grounded answer.
Despite its simplicity, \ours{} achieves strong performance on \oursbench{} and generalizes to open-domain long-video benchmarks, showing that person-centric retrieval is a powerful inductive bias for surveillance understanding.


We evaluate \ours{} on \oursbench{} against off-the-shelf video LMMs and retrieval-augmented baselines.
\ours{} achieves the best overall accuracy (66.0\%) and temporal localization IoU (13.6\%), and ranks first on \oursbench{}$^\texttt{MM}$ accuracy (65.4\%), with the largest gains on the \emph{search} task where person-centric retrieval is most critical.
It further generalizes beyond surveillance to open-domain long-video benchmarks, matching or exceeding state-of-the-art methods while using only half as many input frames.
\ours{} also achieves lower end-to-end latency than all retrieval-augmented baselines (2.6\,s vs.\ 5.2--7.6\,s) and VideoLLaMA3 (3.8\,s), showing that person-centric retrieval reduces the Video LMM frame budget without sacrificing accuracy.



Our main contributions are as follows.
First, we introduce \textbf{\oursbench{}}, the first benchmark for multimodal, temporally grounded video QA in the surveillance domain, covering six subtasks with joint multiple-choice accuracy and temporal localization evaluation under both text-only and multimodal queries.
Second, we present \textbf{\ours{}}, a simple yet strong Video-RAG baseline that combines off-the-shelf person tracking, multimodal embedding, and a VideoLMM into a unified pipeline for forensic search.
Finally, through comprehensive experiments, we show that \ours{} outperforms standard Video LMMs and retrieval-augmented baselines on \oursbench{}, generalizes to open-domain long video benchmarks with competitive performance at half the frame budget, and achieves substantially lower retrieval latency than prior RAG approaches.

\section{Related Work}
\label{sec:related}

\noindent\textbf{Video LMMs.}
Recent LMMs advance video-language reasoning through two main directions: (1) modality integration, where models like Video-LLaVA~\cite{lin2023video} and LLaVA-NeXT-Interleave~\cite{li2024llava} align or interleave visual tokens with text for multi-frame understanding; and (2) scalability, with VideoLLaMA3~\cite{zhang2025videollama} applying token compression for long videos, while InternVL~\cite{chen2024expanding} and Qwen2.5-VL~\cite{bai2025qwen25vl} leverage large-scale multimodal data and powerful language backbones. Despite these advances, most Video LMMs process the full video end-to-end without external knowledge grounding, which limits performance on long-horizon QA tasks where relevant evidence is sparse.

\noindent\textbf{Retrieval-Augmented Video Understanding.}
VideoRAG systems combine retrieval from large-scale video corpora with generative models to support long-form video QA. Recent advances include visually-aligned retrieval, graph-based grounding, memory-enhanced retrieval, and adaptive temporal search~\cite{jeong2025videorag, ren2025videorag, luo2024video, ye2025re, sagare2024videorag, yuan2025memory, mao2025multi}. In the surveillance domain, video anomaly detection (VAD) methods have adopted language-guided and retrieval-augmented techniques for identifying rare events, including training-free LLM-based scoring, spatiotemporal graph reasoning, and verbalized learning~\cite{zanella2024harnessing, shao2025eventvad, zhang2025holmes, ye2025vera}. However, existing VideoRAG systems are designed for general-purpose QA and lack fine-grained temporal localization, while VAD methods target classification or anomaly scoring rather than interactive, multimodal question answering.

\noindent\textbf{Multimodal Retrieval.}
While cross-modal retrieval focuses on single-modality mappings like image-to-text (e.g., CLIP~\cite{clip}), multimodal retrieval enables flexible searches across mixed modality pairs~\cite{gcl,vista}. Systems such as VISTA~\cite{vista} and GCL~\cite{gcl} allow queries and targets to include images, text, or both, supporting unified retrieval across heterogeneous inputs. Despite this flexibility, multimodal retrieval remains underexplored for forensic search, where combining image and text queries is crucial for identifying specific individuals.

\noindent\textbf{Benchmarks.}
General-purpose long video benchmarks, such as InfiniBench~\cite{ataallah2024infinibench}, LoVR~\cite{cai2025lovr}, and LongerVideos~\cite{ren2025videorag}, support long-form retrieval but lack detailed temporal annotations and multimodal query support. Domain-specific benchmarks like TUMTraffic-VideoQA~\cite{zhou2025tumtraffic}, SurveillanceVQA-589K~\cite{liu2025surveillancevqa} and SmartHome-Bench~\cite{zhao2025smarthome} address traffic, surveillance, and smart home scenarios but restrict queries to a single modality. Event-focused datasets like MomentSeeker~\cite{yuan2025momentseeker} emphasize temporal retrieval but target single events rather than complex forensic contexts. \oursbench{} is the first benchmark to jointly evaluate multiple-choice accuracy and temporal localization under both text-only and multimodal query conditions in the surveillance domain.

\section{\oursbench{}: Benchmarking Grounded Multimodal Video Understanding}
\label{sec:data}

We construct the \oursbench{} benchmark to evaluate the ability of LMMs to understand long videos, ground people and moments of interest, and answer questions based on the retrieved evidence. 

\subsection{Benchmark Design}
The benchmark differs from existing long video benchmarks by introducing two unique challenges to the models.

\paragraph{Joint answer and localization.}
We augment each question-answer pair with time ranges of grounded evidence that supports the answer, and require models to jointly output its answer with the associated timestamps. Specifically, we construct the dataset as $\mathcal{D} = \{(V, Q, A, T)\}$, where $T$ can be one or multiple intervals $T = \{(T_s, T_e)\}$ that contain sufficient and necessary information from video $V$ to predict the correct answer $A$ of question $Q$.
While such time annotations are used in some existing benchmarks (e.g., Charades-STA~\cite{gao2017tall}, VideoSIAH~\cite{yang2025longvt}) benchmarks, they are often limited to a single interval or a list of non-exhaustive keyframes per question, and usually do not evaluate localization and question answering tasks jointly.

\paragraph{Multimodal queries.}
In addition to text-only questions, \oursbench{} includes \emph{multimodal} queries with supplementary images to the question. Concretely, each multimodal query is represented as $Q = (Q_I, Q_T)$ where $Q_I$ is an image and $Q_T$ is a question that \emph{refers to} the query image (e.g. ``When did \emph{this person} enter the building?'').
This mirrors practical scenarios in surveillance analysis, where a snapshot of a person of interest is provided as reference to enable tasks such as identifying when and where the individual appears, or what activities they participate in; answering such questions require LMMs to simultaneously understand the video frames, the reference image and the question interleaved in the same multimodal input sequence, a capability rarely examined in prior video benchmarks.

\begin{figure}[t]
    \centering
    \includegraphics[width=0.70\linewidth]{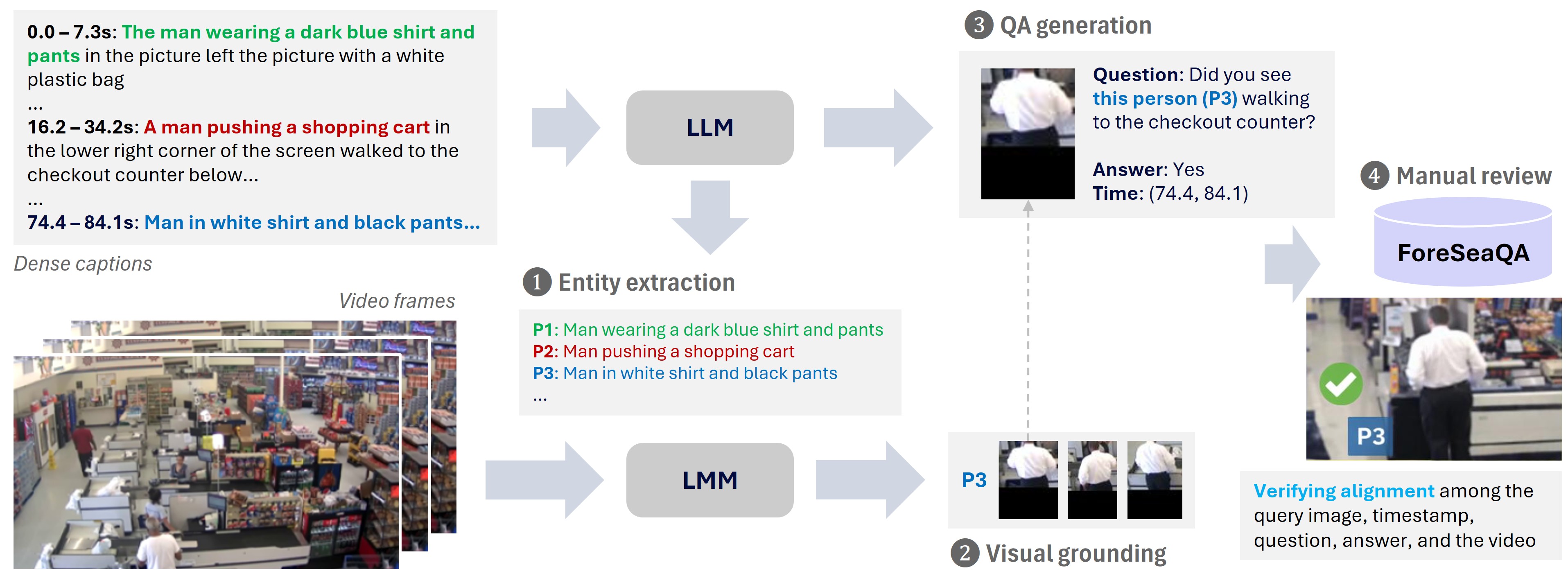}
    \caption{\textbf{\oursbench{} Data Engine}. We use text-only and multimodal LLMs to \circled{1} extract person entities from dense video captions, \circled{2} visually ground each entity to create query image crops, and \circled{3} generate multimodal QA pairs with timestamps. All generated QA samples and query images are \circled{4} reviewed by human workers for correctness.}
    \label{fig:data-engine}
\end{figure}

\subsection{Data Engine}
\label{subsec:data_engine}

We use videos from the UCF-Crime dataset~\cite{sultani2018real} and a semi-automated data engine to generate \emph{temporally grounded} and \emph{multimodal} video QA from dense captions, as illustrated in Figure~\ref{fig:data-engine}. 
The engine has 4 stages:

\noindent{\textbf{\circled{1} Entity extraction:}}
A text-only LLM\footnote{Qwen3-32B~\cite{yang2025qwen3} for QA generation and Qwen2.5-VL-32B~\cite{bai2025qwen25vl} for spatial grounding.} parses dense UCA~\cite{yuan2023surveillance} captions to extract human entity references (e.g., ``man in white shirt''). Multiple references to the same individual are grouped, creating a list of timestamps per person.

\noindent{\textbf{\circled{2} Visual grounding:}}
We use a LMM to ground the extracted entities.
For each timestamp from \circled{1}, we sample 8 frames uniformly within the annotated timestamp and ask the model to predict bounding boxes for the referred person. We then crop the bounding boxes and prompt the LMM again to verify the person's presence to prevent hallucinated coordinates.
The crops of person entities are used as query images in multimodal questions of \oursbench{}.

\noindent{\textbf{\circled{3} Grounded QA generation:}}
We then use the text LLM to generate candidate QA pairs from the captions.\footnote{Generation prompts per question type are provided in the supplemental material.}
\oursbench{} includes questions from 6 subtasks: \emph{search} (SE), \emph{activity} (AC), \emph{event} (EV), \emph{temporal} (TM), \emph{counting} (CT), and \emph{anomaly} (AN).
Among these, search, activity, event and temporal questions are \emph{person-specific} and are generated for each person entity; counting and anomaly questions are \emph{global} and generated for the entire video.
For each answer, the LLM assigns temporal groundings by selecting time ranges from the timestamp lists obtained in \circled{1}.
To create multimodal questions in person-specific tasks, we rephrase the question to refer indirectly to the grounded entity images from \circled{2} (e.g., using ``the person in the photo'' instead of ``the man in the white shirt'').

\noindent{\textbf{\circled{4} Manual verification:}}
We manually validate all generated QA pairs by 2 stages.
1) Human reviewers verified the semantic alignment of all QA pairs against the video, aggressively removing 50\% of the initial generations. 
2) We used VideoLLM\footnote{We employed Qwen3-VL-32B\cite{Qwen3-VL} to verify between video and generated QA pairs.} 
with UCA captions to detect misalignment again, and then correct every misaligned QA pair, query image and timestamp manually.

\begin{figure*}[t]
    \centering
    \begin{subfigure}[b]{0.22\textwidth}
        \centering
        \includegraphics[width=\linewidth]{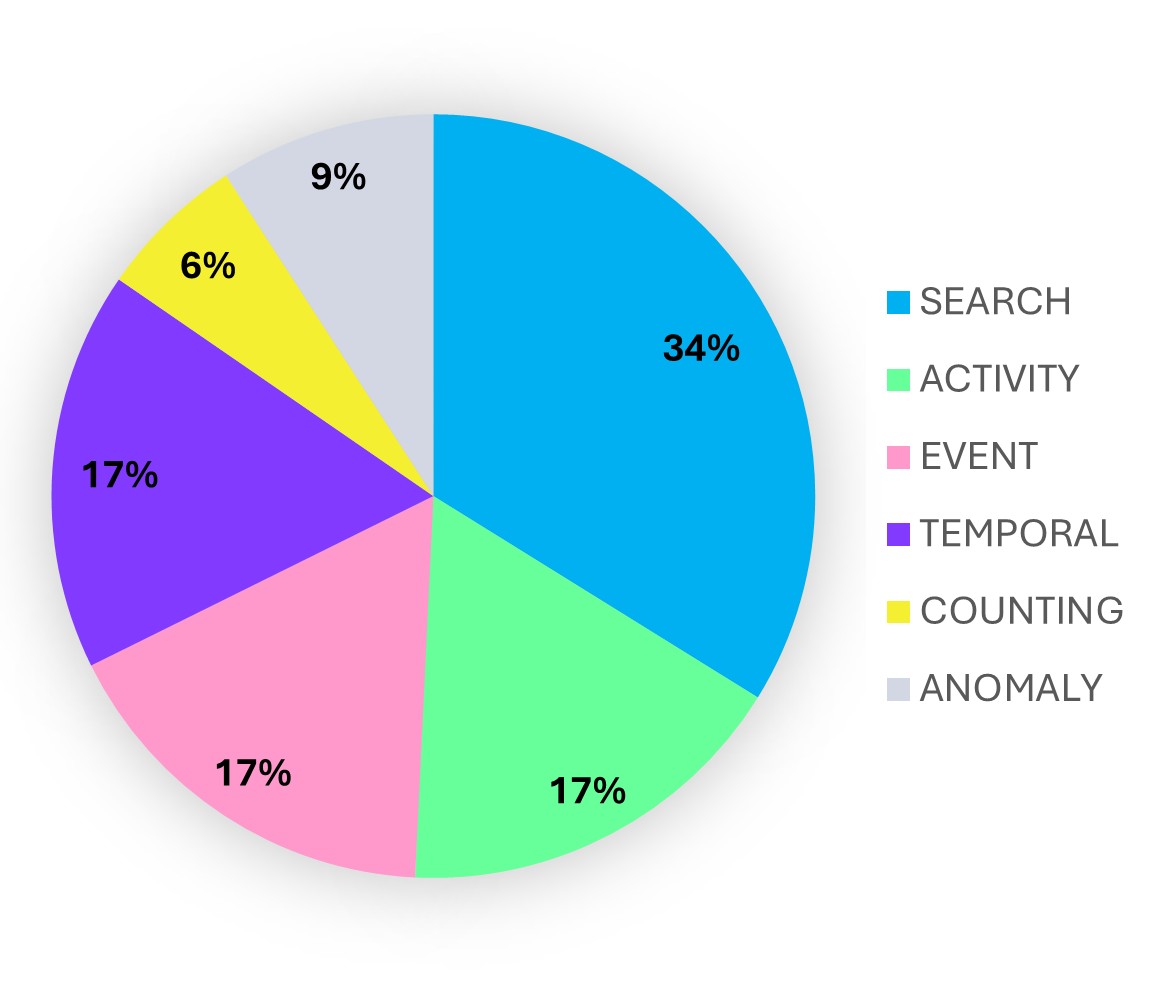}
        \caption{}
        \label{fig:bench-stats-pie}
    \end{subfigure}\hfill
    \begin{subfigure}[b]{0.22\textwidth}
        \centering
        \includegraphics[width=\linewidth]{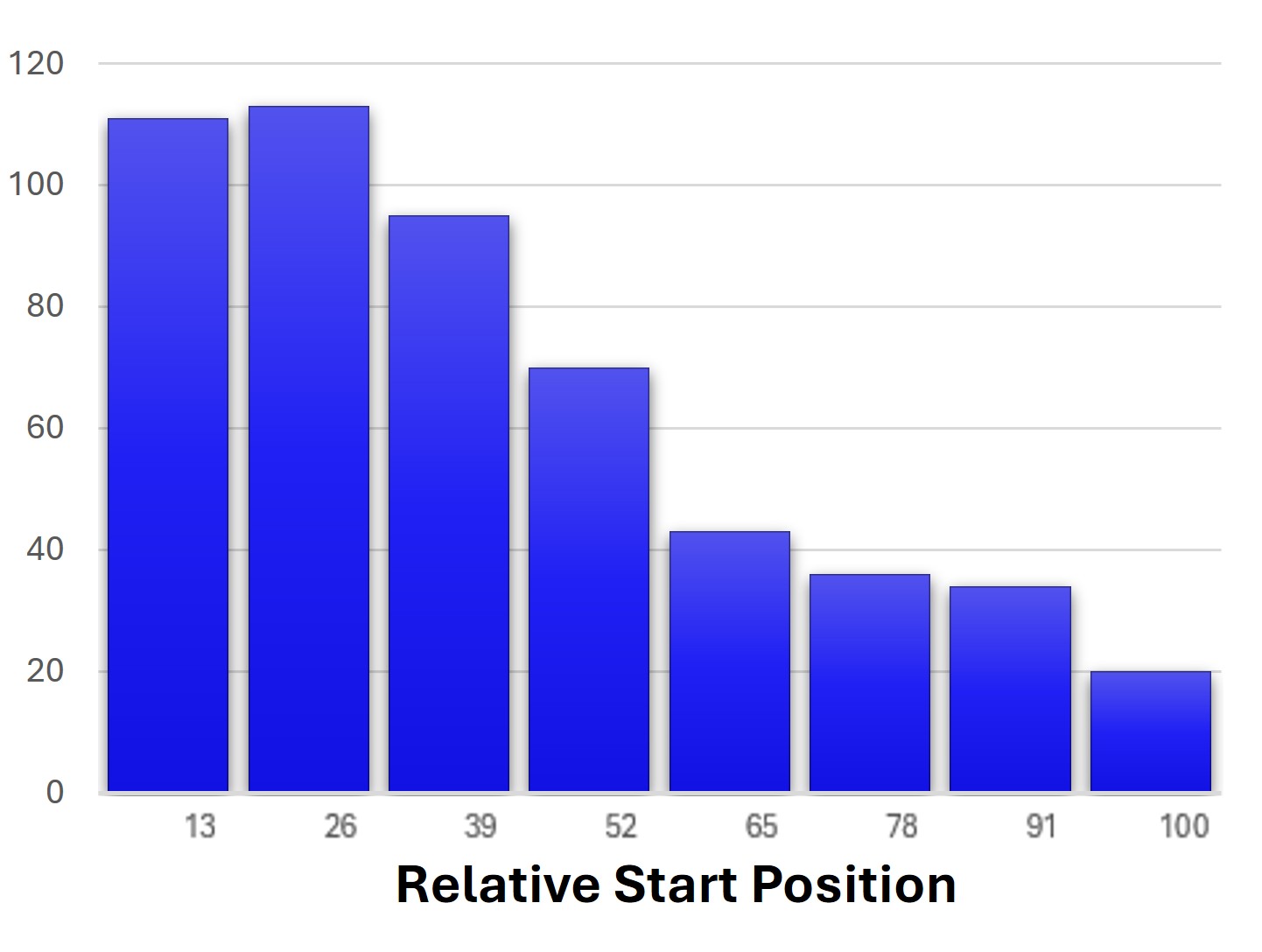}
        \caption{}
        \label{fig:bench-stats-startratio}
    \end{subfigure}\hfill
    \begin{subfigure}[b]{0.20\textwidth}
        \centering        
       \resizebox{\linewidth}{!}{
       \begin{tabular}{l r}
            \toprule
            \textbf{Statistics} & \textbf{Length (sec)} \\
            \midrule
            Min     & 77.35 \\
            Max     & 2112.88 \\
            Mean    & 352.94 \\
            Median  & 262.83 \\
            Std     & 374.60 \\
            25th    & 142.95 \\
            75th    & 399.16 \\
            \bottomrule
           \end{tabular}
        }
        \caption{}
        \label{tab:video-duration-stats}
    \end{subfigure}\hfill
    \begin{subfigure}[b]{0.28\textwidth}
        \centering
        \resizebox{\linewidth}{!}{
        \begin{tblr}{
          colspec = {l c c c},
          rowsep = 1pt,
          rows   = {abovesep=0.5pt, belowsep=0.5pt},
          cell{2}{1} = {c=4}{},
          cell{6}{1} = {c=4}{},
          cell{9}{1} = {c=4}{},
          hline{1,13} = {-}{0.08em},
          hline{2,6,9} = {-}{},
        }
        \textbf{Benchmark} & \textbf{Tasks} & \textbf{T\_{ann}} & \textbf{MMq} \\
        \textit{Comprehensive} & & & \\
        LongVid\cite{wu2024longvideobench} & MC & \wrong & \wrong \\
        LVBench\cite{wang2025lvbench} & MC & \correct & \wrong \\
        Vid-MME\cite{fu2025videomme} & MC & \wrong & \wrong \\
        \textit{Temporal retrieval} & & & \\
        ICQ-High\cite{zhang2024localizing} & TG & \correct & \correct \\
        MSeeker\cite{yuan2025momentseeker} & TG & \correct & \correct \\
        \textit{Surveillance domain} & & & \\
        TUMT-VQA\cite{zhou2025tumtraffic} & MC, STG & \correct & \wrong \\
        SVQA-589K\cite{liu2025surveillancevqa} & OE & \correct & \wrong \\
        \oursbench{}~(ours) & MC, TG & \correct & \correct \\
        \end{tblr}
        }
        \caption{}
        \label{fig:bench-stats-table}
    \end{subfigure}
    
    \caption{Statistics of \oursbench{} benchmark. 
    (a) Task distribution by question. 
    (b) Relative start position of ground-truth time ranges.
    (c) Statistics of video duration.
    (d) Comparison of benchmarks. Tasks: MC=multiple-choice, OE=open-ended, TG=temporal grounding, STG=spatiotemporal grounding. T\_{ann}= Temporal annotation, MMq =Multimodal query. }
    \label{fig:bench-stats}
\end{figure*}

\subsection{Benchmark Details}
Following the procedure described in Section~\ref{subsec:data_engine}, we construct the final \oursbench{} benchmark, which comprises 1,041 curated questions. Figure~\ref{fig:bench-stats} summarizes key dataset statistics, including the subtask distribution (Figure~\ref{fig:bench-stats}a), the relative starting positions of annotated temporal windows (Figure~\ref{fig:bench-stats}b), and video-length statistics (Figure~\ref{fig:bench-stats}c).
The benchmark spans a wide range of video durations and temporal intervals. The starting points of the annotated time ranges vary substantially across questions, demonstrating that temporal grounding in \oursbench{} cannot be solved by heuristics that focus only on early or late portions of the video.
While the benchmark places particular emphasis on \emph{search} questions---reflecting their role as a foundation for more advanced temporal reasoning tasks---it also provides balanced coverage of \emph{activity}, \emph{event}, \emph{temporal}, and global tasks such as \emph{counting} and \emph{anomaly detection}. This diversity ensures that models are evaluated across a broad spectrum of forensic video understanding capabilities.



\section{Method}
\label{sec:method}
We present our \ours{}, a novel videoRAG framework designed for multimodal queries.
In Sec. \ref{subsec:arch}, we describe the overall system architecture about how we build the searchable database, and how our model provides answers for multimodal surveillance queries.
In Sec. \ref{subsec:emb}, we describe the multimodal encoder in detail.
We explain how it encodes visual and textual inputs into a unified embedding space, how these embeddings are stored in the database, and how they are later used during retrieval.
Finally, in Sec. \ref{subsec:VidLLM}, we explain how the VideoLLM stage answers user queries.

\begin{figure*}[t]
  \centering
   \includegraphics[width=0.95\linewidth]{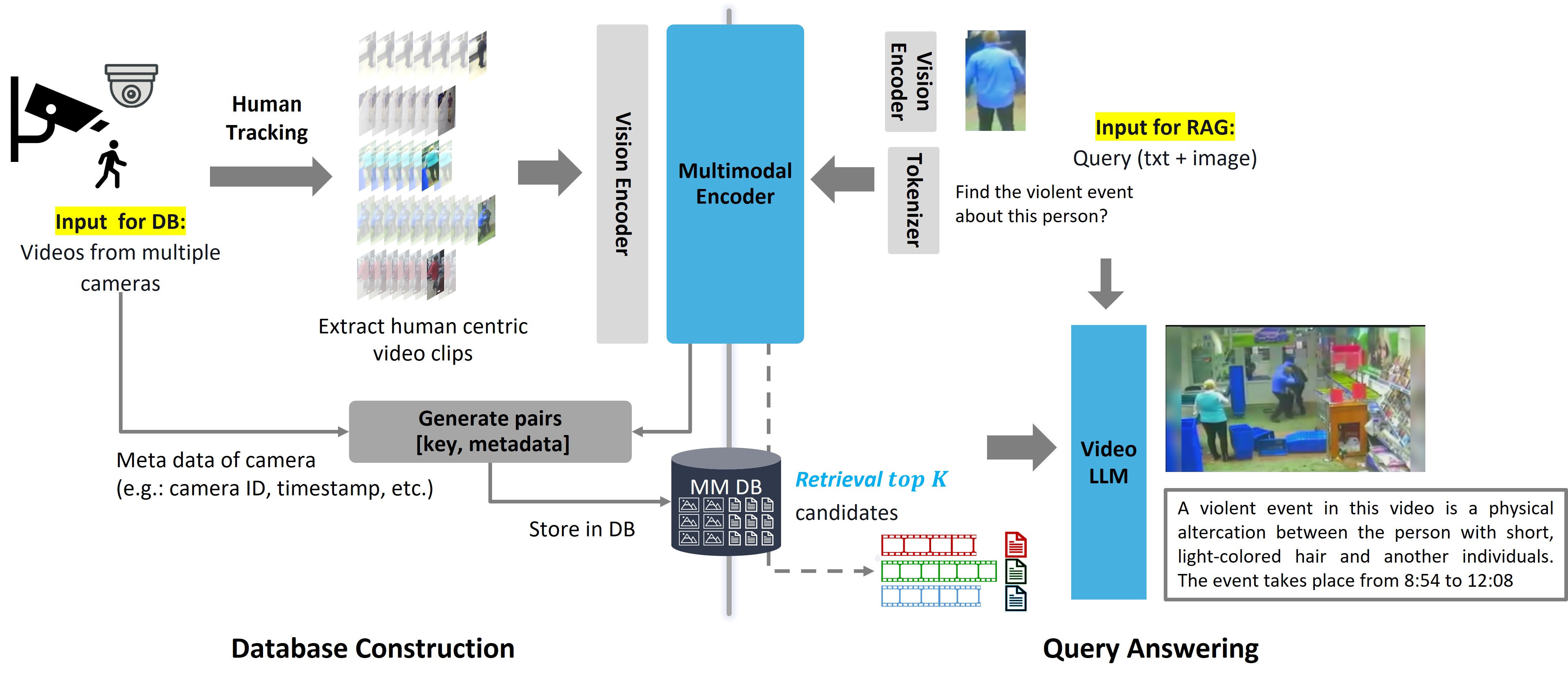}

   \caption{
   \textbf{Overview of \ours{} Pipeline.} \ours{} consists of two main components: 
   (1) Video Database Construction—a multimodal encoder embeds short video clips from the human tracking module and pairs them with metadata; 
   (2) Query Answering—retrieves candidate videos from the database using a multimodal query and generates answers based on the retrieved content
   }
   \label{fig:arch}
\end{figure*}

\subsection{Overall Architecture}
\label{subsec:arch}
The overall architecture of the proposed system is illustrated in Figure~\ref{fig:arch}. 
The pipeline consists of two stages: (i) video database construction and (ii) query answering with VideoLMM reasoning.

\noindent\textbf{Video Database Construction:} The system begins by collecting raw video recordings $D$ from multiple cameras. 
A human tracking module processes these videos to extract only relevant frames, and $D$ is segmented into short clips according to the tracking results. 
Each segment is then cropped using the corresponding bounding box coordinates to produce human-centric video clips $C=\{c_1,\dots,c_j\}$. 
Each clip $c_j$ is fed into the multimodal encoder (detailed in \ref{subsec:emb}) to generate a database embedding vector $\mathbf{e}_j^d$. 
This vector $\mathbf{e}_j^d$, which captures the semantic content of the clip, is stored in a multimodal database together with relevant metadata\footnote{We use camera ID, timestamp, and bounding box coordinates.} to enable efficient retrieval.

\noindent\textbf{Query Answering.}
The system supports various query formats, including text-only queries ($q_t$) and image–text queries ($q_{i\!t}$). 
Given a query, the same multimodal encoder is used to generate a unified query embedding $\mathbf{e}^q$. 
This vector is matched against the database to retrieve the top-$K$ candidate embeddings $\{\mathbf{e}_j^d\}$. 
The corresponding top$K$ candidate clips are then concatenated and provided as input to a VideoLMM, along with the original query and augmented information (such as bounding box coordinates), to produce a summary of key events and a temporally grounded answer with linked visual evidence.

\subsection{Multimodal Embedding}
\label{subsec:emb}

\begin{figure}[t]
  \centering
    \includegraphics[width=0.7 \linewidth]{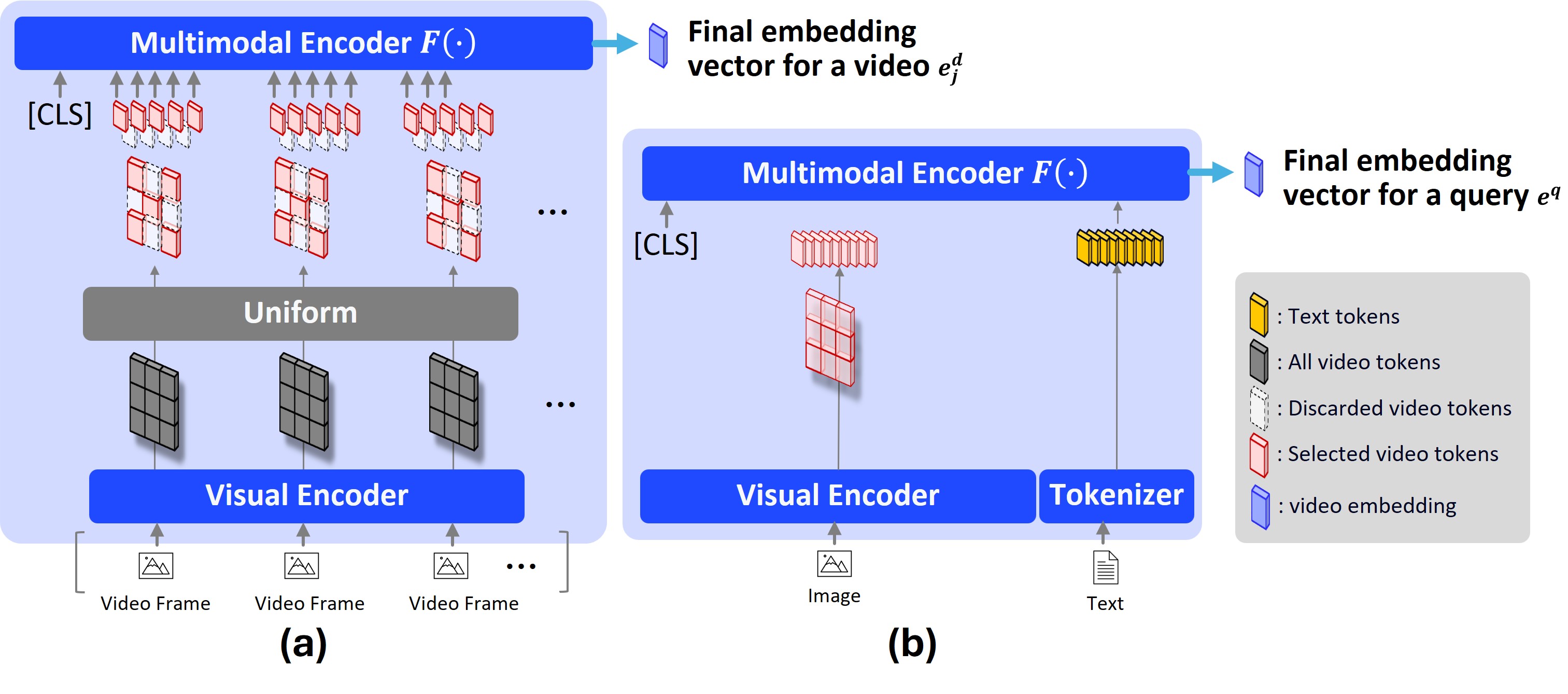}
    \vspace{-3mm}

   \caption{Multimodal encoder produces (a) a video embedding from multiple frames and (b) a query embedding from text or image-text inputs}
   \label{fig:mmEnc}
\end{figure}

We build both the retrieval index and the query embeddings using a publicly available multimodal encoder introduced in \cite{vista,gcl}, as shown in Figure~\ref{fig:mmEnc}.

\noindent\textbf{Video embedding: }
As shown in \ref{fig:mmEnc} (a), for each clip $c_j$ from tracking module, we obtain frames $C_j=\{f_{j,k}\}_{k=1}^{m_j}$ and compute frame-level visual tokens $\mathbf{x}_j^d=\{\mathbf{x}_{j,1}^d,\dots,\mathbf{x}_{j,m_j}^d\}$ with the visual encoder.
Here, $m_j$ denotes the total number of frames for the $j$-th clip.
To ensure a consistent number of input tokens for the MMEnc (equivalent to that of a single image input), we apply uniform sampling $S(\cdot)$\footnote{The sampling rate adapts to the length of video clips, so the resulting number of tokens matches that of a single-image input. Because the human tracking clips have variable lengths, and uniform sampling provides a fixed-size representation.}.
We feed the sampled tokens to MMEnc and use the [CLS] output as the database vector $\mathbf{e}_i^d\in\mathbb{R}^p$:
\begin{equation}
\mathbf{e}_j^d = \mathrm{MMEnc}\,\!\big([\mathbf{x}_{CLS},S(\mathbf{x}_j^d)]\big)
\end{equation}

\noindent\textbf{Query embedding: }
We use the same MMEnc for all query formats: text ($q_t$) and image+text($q_{i\!t}$) as shown in \ref{fig:mmEnc} (b).
The text query is tokenized into $x_t^q$, and an image is encoded by the visual encoder into $x_i^q$.
For a text-only query, $x^q_t$ serves as the input $x^q$ to the MMEnc. 
For an image-text query, the visual features $x^q_i$ are passed through a projection layer to match the dimension of $x^q_t$. 
Both sets of features are then concatenated to form the final input $x^q$.
In all cases, the [CLS] output gives the query vector $e^q\in\mathbb{R}^p$:
\begin{equation}
\mathbf{e}^q = \mathrm{MMEnc}\,([\mathbf{x}_{CLS},\mathbf{x}^q]), \quad \text{where } \mathbf{x}^q \in \{\mathbf{x}_t^q,\ [\mathbf{x}_i^q;\,\mathbf{x}_t^q]\} 
\end{equation}
Most existing approaches perform retrieval in a text-only space, converting all modalities (video frames, ASR transcripts, etc.) into text. This "unimodal projection" inherently leads to information loss. In contrast, ForeSea performs retrieval directly within a unified multimodal embedding space. This approach not only avoids information loss and yields superior accuracy, but it also ensures that semantically relevant instances are retrieved regardless of the query modality, enabling a truly flexible and scalable multimodal search.

\subsection{Response Generation from Retrieval Results}
\label{subsec:VidLLM}

Following the retrieval stage, we obtain a set of top-$K$ candidate video clips, each associated with precise spatio-temporal metadata: a start and end timestamp ($T_s, T_e$) and bounding box coordinates ($bbox$). 
To prepare input for the videoLMM, we extract frames from each candidate clip at source resolution and draw $bbox$ on every frame.
This augmentation explicitly directs the model's attention to the people.
We guide the VideoLMM's output by providing a system prompt engineered to solicit two key pieces of information: (1) a concise summary of the events occurring within the spatio-temporal window, and (2) a list of precise timestamps for any key events observed.

\section{Experiments}
\label{sec:exp}

In this section, we evaluate a range of existing Video LMMs and retrieval-augmented baselines on \oursbench{}, and present \ours{} as a strong baseline for multimodal forensic search.
We further demonstrate that \ours{} generalizes to open-domain long video benchmarks.
Sec.~\ref{subsec:AI-QAexp} presents main results on \oursbench{} under both text-only and multimodal query conditions.
Sec.~\ref{subsec:ablation} and \ref{subsec:ablations_vlm} ablate the key design choices of \ours{}.
Sec.~\ref{subsec:latency} compares efficiency across methods.
Sec.~\ref{subsec:vidBench_exp} evaluates \ours{} on VideoMME and MLVU to assess generalization beyond the surveillance domain.

\subsection{Experimental Setup}
\label{subsec:setup}

\noindent\textbf{Evaluation Protocols and Metrics.}
All evaluations on \oursbench{} are conducted under two query conditions: \emph{text-only} (\oursbench{}$^\texttt{Text}$) and \emph{multimodal} image+text (\oursbench{}$^\texttt{MM}$), as described in Sec.~\ref{sec:data}.
We report \emph{accuracy} (percentage of correctly answered multiple-choice questions) and temporal localization \emph{IoU} (intersection-over-union between the predicted and ground-truth time intervals, averaged over all questions) as the two primary metrics.

\noindent\textbf{Models.}
We evaluate a diverse set of Video LMMs and retrieval-augmented baselines on \oursbench{}.
For Video LMMs, we include LLaVA-OneVision~\cite{li2024llavaonevision}, GLM-4.1V-Thinking~\cite{hong2025glm}, InternVL3~\cite{zhu2025internvl3}, Qwen2.5-VL~\cite{bai2025qwen25vl}, and VideoLLaMA3~\cite{zhang2025videollama}, spanning model sizes from 2B to 72B parameters.
For retrieval-augmented baselines, we include VideoRAG~\cite{luo2024video} and T$^\ast$~\cite{ye2025re}.
We also evaluate our proposed \ours{} (Sec.~\ref{sec:method}).

\noindent\textbf{Implementation Details.}
\ours{} uses ByteTrack~\cite{zhang2022bytetrack} with a YOLO-based~\cite{yolov5} detector to segment long videos into person-centric clips, which are indexed using a GCL-trained~\cite{gcl} multimodal encoder following VISTA~\cite{vista}.
During retrieval, it selects the top-$K$ ($K=3$) most relevant clips and passes them to VideoLLaMA3~\cite{zhang2025videollama} for answer generation.

\subsection{Results on \oursbench{}}
\label{subsec:AI-QAexp}

\begin{table}[t]
\centering
\footnotesize
\caption{\textbf{Performance comparison on \oursbench{}.} \oursbench{}$^\texttt{MM}$ and \oursbench{}$^\texttt{Text}$ denote multimodal (image+text) and text-only query; \oursbench{} reports their average.}
\label{tab:bench_ours}
\resizebox{0.8\linewidth}{!}{%
\setlength{\tabcolsep}{5pt}
\begin{tabular}{l c cc cc cc}
\toprule
\multirow{2}{*}{\textbf{Model}} & \multirow{2}{*}{\textbf{Params}} &
\multicolumn{2}{c}{\textbf{\oursbench{}$^\texttt{MM}$}} &
\multicolumn{2}{c}{\textbf{\oursbench{}$^\texttt{Text}$}} &
\multicolumn{2}{c}{\textbf{\oursbench{}}} \\
\cmidrule(lr){3-4}\cmidrule(lr){5-6}\cmidrule(lr){7-8}
& & Acc & IoU & Acc & IoU & Acc & IoU \\
\midrule
\multicolumn{8}{c}{\textit{Video LMMs}} \\
\midrule
\multirow{1}{*}{LLaVA-OneVision~\cite{li2024llavaonevision}} & 7B  & 56.1 & 10.4 & 58.5 & 7.7  & 57.3 & 9.0 \\
\multirow{1}{*}{GLM-4.1V-Thinking~\cite{hong2025glm}} & 9B  & 57.4 & 10.0 & 55.1 & 8.4  & 56.2 & 9.2 \\
InternVL3~\cite{zhu2025internvl3} & 2B  & 38.3 & 9.8  & 34.1 & 7.1  & 36.2 & 8.4 \\
& 8B  & 61.3 & 10.2 & 63.4 & 9.9  & 62.3 & 10.0 \\
& 9B  & \underline{62.3} & 11.5 & 62.7 & 8.8  & 62.5 & 10.2 \\
Qwen2.5-VL~\cite{bai2025qwen25vl} & 7B  & 58.9 & 8.1  & 59.0 & 7.5  & 58.9 & 7.8 \\
& 72B & 60.0 & \textbf{15.3} & 61.4 & 10.1 & 60.7 & 12.7 \\
VideoLLaMA3~\cite{zhang2025videollama} & 7B & 61.6 & 10.9 & \textbf{67.7} & \textbf{15.5} & \underline{64.6} & \underline{13.2} \\
\midrule
\multicolumn{8}{c}{\textit{Retrieval-augmented}} \\
\midrule
\multirow{1}{*}{VideoRAG~\cite{luo2024video}} & 7B & 61.9 & 2.8  & 63.8 & 4.3  & 62.9 & 3.5 \\
\multirow{1}{*}{T$^\ast$~\cite{ye2025re}} & 7B & 41.1 & 4.9  & 48.4 & 4.2  & 44.8 & 4.6 \\
\rowcolor{blue!10} \multirow{1}{*}{\textbf{\ours{} (Ours)}} & 7B  & \textbf{65.4} & \underline{13.8} & \underline{66.7} & \underline{13.3} & \textbf{66.0} & \textbf{13.6} \\
\bottomrule
\end{tabular}%
}
\end{table}


\paragraph{Main results.}
We evaluate all models on \oursbench{}$^\texttt{Text}$ and \oursbench{}$^\texttt{MM}$ and calculate their average as the final \oursbench{} scores; results are reported in Table~\ref{tab:bench_ours}.
We highlight three key observations on the benchmark:

\noindent\textbf{Temporal localization is the primary challenge.}
Despite achieving reasonable multiple-choice accuracy, all Video LMMs produce low temporal localization IoU (7--16\%), indicating that correct answers are often inferred from global video context rather than grounded evidence.
Retrieval-augmented baselines (VideoRAG, T$^\ast$) fare even worse on IoU (2.8--4.9\%), despite comparable or lower accuracy—suggesting that their retrieval strategies do not produce temporally precise evidence.
In contrast, \ours{} achieves substantially higher IoU (13.6\%), demonstrating that person-centric retrieval is a strong inductive bias for temporal grounding in surveillance videos.

\noindent\textbf{Multimodal queries expose a gap in existing Video LMMs.}
\oursbench{}$^\texttt{MM}$ is consistently harder than \oursbench{}$^\texttt{Text}$ for most models, with accuracy dropping by up to 6 points (e.g., VideoLLaMA3: 67.7\%$\to$61.6\%).
This suggests that current Video LMMs struggle to jointly reason over a reference image and a long video—a capability central to forensic search.
\ours{} is more robust to this shift: it maintains accuracy above 65\% on both \oursbench{}$^\texttt{Text}$ (66.7\%) and \oursbench{}$^\texttt{MM}$ (65.4\%), while no other method does.

\noindent\textbf{Accuracy--localization tradeoff.}
\ours{} achieves the best overall accuracy (66.0\%) and IoU (13.6\%) among all retrieval-augmented methods, and ranks first on \oursbench{}$^\texttt{MM}$ accuracy (65.4\%) across all evaluated models.
Notably, \ours{} outperforms all Video LMMs on \oursbench{}$^\texttt{MM}$ accuracy while using only 7B parameters, demonstrating that person-centric retrieval provides a meaningful advantage over dense video processing for multimodal forensic queries.


\begin{figure*}[t]
\centering
\includegraphics[width=0.9\linewidth]{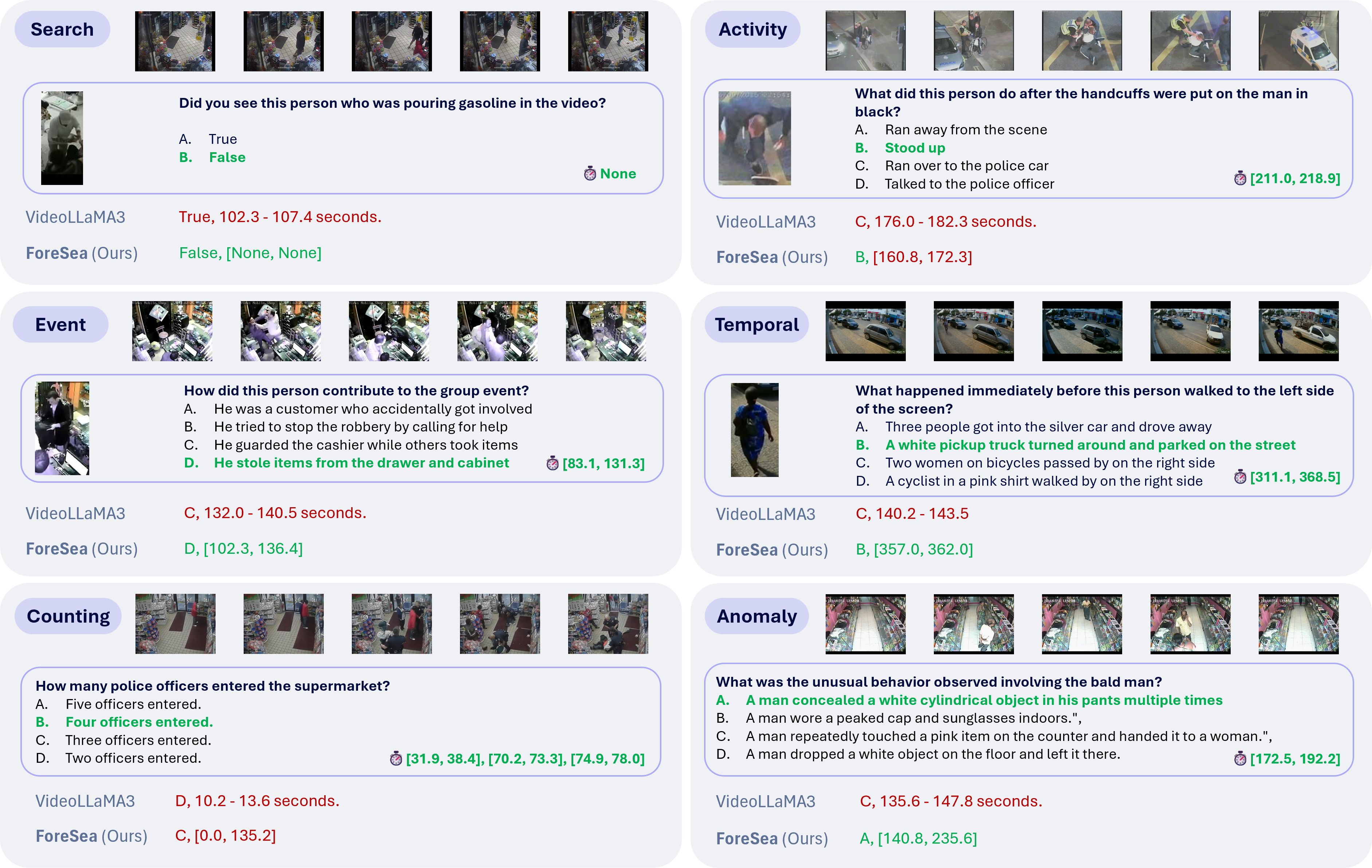}
\caption{\textbf{Qualitative examples of \ours{} and VideoLLaMA3 on \oursbench{}.} Ground-truth answers are highlighted in green. Model answers are highlighted in \textcolor{Green}{green} if correct (multiple-choice) or have nonzero IoU (temporal grounding), and \textcolor{Red}{red} if wrong.}
\label{fig:qualitative}
\end{figure*}

\paragraph{Qualitative examples.}
Figure~\ref{fig:qualitative} shows a qualitative comparison between \ours{} and VideoLLaMA3 on samples of different tasks of \oursbench{}. In \emph{event}, \emph{temporal} and \emph{anomaly} examples, \ours{} correctly identifies the time intervals containing the relevant information and answers correctly, while VideoLLaMA3 fails to localize the evidence and produces wrong answers. In the \emph{search} example where a nonexistent moment is queried, \ours{} correctly identifies the absence of evidence, while VideoLLaMA3 hallucinates a false temporal interval. In the \emph{activity} example, both models fail to localize the moment of interest, but \ours{} still answers correctly by leveraging the retrieved clips. The hardest of all is the \emph{counting} task, where both models under-count the occurrences and fail to follow the output format by providing a list of time intervals, suggesting that counting-based video QA remains a challenging open problem that requires more sophisticated retrieval and reasoning strategies.






\subsection{Ablation Studies of ForeSea Configurations}
\label{subsec:ablation}

We ablate the key design choices of \ours{} on \oursbench{}$^\texttt{MM}$ in Table~\ref{tab:ablation}, including VideoLLaMA3-7B as the no-retrieval baseline.
After retrieval, each track is passed to the Video LMM together with optional spatial grounding signals: \textbf{Crop} crops the video frames to the tracked bounding box; \textbf{Overlay} draws the bounding box on the original (uncropped) frames; \textbf{Coords} appends the bounding box coordinates as text in the prompt.
Top\,$K$ controls how many retrieved tracks are concatenated as Video LMM input.

\noindent\textbf{Person-centric retrieval alone outperforms direct video processing.}
Even without any spatial grounding (no Crop, no Overlay, no Coords), \ours{} with $K{=}3$ already surpasses VideoLLaMA3-7B on both accuracy (64.6\% vs.\ 61.6\%) and temporal IoU (12.5\% vs.\ 10.9\%).
This confirms that focusing the Video LMM on a small set of person-centric clips, rather than the full video, is itself a strong inductive bias for forensic search, even before any explicit spatial information is provided.

\begin{table}[t]
\centering
\small
\caption{Ablation study on \oursbench{}$^\texttt{MM}$. The \colorbox{blue!10}{highlighted row} indicates the default configuration used in the main results. 
$^\ast$ denotes top $K$ retrieval of \emph{global frames} rather than person-centric clips.
}

\label{tab:ablation}
\setlength{\tabcolsep}{4pt}
\renewcommand{\arraystretch}{1.1} 
\resizebox{0.9\textwidth}{!}{%
\begin{tabular}{lcccc|ccccc|ccccc}
\hline
\multirow{2}{*}{Model} & \multicolumn{4}{c|}{Setup} & \multicolumn{5}{c|}{Multi-choice Acc. (\%)} & \multicolumn{5}{c}{Temporal Loc. IoU (\%)}\\
\cline{2-5} \cline{6-10} \cline{11-15} 
& Crop & Overlay & Coords & Top\,$K$ & Search & Act. & Event & Temp. & Avg & Search & Act. & Event & Temp. & Avg\\
\hline
VideoLLaMA3-7B & - & - & - & -
& 49.5 & \textbf{61.0} & 83.0 & 53.0 & 61.6
& 10.7 & 15.0 & 9.8 & 8.2 & 10.9 \\
\hline
\multirow{8}{*}{\ours{}}
& \wrong & \wrong & \wrong & 3
& 58.5 & 58.0 & \underline{87.0} & 55.0 & 64.6
& 14.8 & 12.8 & 12.5 & 9.8 & 12.5 \\
& \correct & \wrong & \wrong & 3
& 53.0 & 54.0 & 82.0 & \underline{56.0} & 61.3
& 14.0 & 11.3 & 14.2 & 10.4 & 12.5 \\
& \wrong & \correct & \wrong & 3
& 60.0 & 56.0 & 85.0 & \underline{56.0} & 64.3
& 15.0 & 12.5 & 13.8 & 11.5 & 13.2 \\
& \wrong & \correct & \correct & 3
& \textbf{61.0} & 59.0 & 85.0 & 53.0 & 64.5
& 15.0 & 10.1 & 13.9 & 9.2 & 12.1 \\
& \cellcolor{blue!10}\wrong & \cellcolor{blue!10}\wrong & \cellcolor{blue!10}\correct & \cellcolor{blue!10}3
& \cellcolor{blue!10}\underline{60.5} & \cellcolor{blue!10}\underline{60.0} & \cellcolor{blue!10}85.0 & \cellcolor{blue!10}\underline{56.0} & \cellcolor{blue!10}\underline{65.4}
& \cellcolor{blue!10}\textbf{17.6} & \cellcolor{blue!10}14.1 & \cellcolor{blue!10}15.4 & \cellcolor{blue!10}8.2 & \cellcolor{blue!10}13.8 \\
& \wrong & \wrong & \correct & 5
& 59.5 & \textbf{61.0} & \textbf{88.0} & \textbf{57.0} & \textbf{66.4}
& 15.8 & 10.6 & 13.2 & 8.9 & 12.1 \\
\hline
\ours{}-Global & - & - & - & 64*
& 57.5 & 58.0 & \textbf{88.0} & \textbf{57.0} & 65.1
& 14.1 & \textbf{18.4} & \underline{19.3} & \underline{12.7} & \underline{16.1} \\
\ours{}-Hybrid & \wrong & \wrong & \correct & 2/24*
& 59.5 & 59.0 & \underline{87.0} & 53.0 & 64.6
& \underline{16.9} & \underline{15.1} & \textbf{20.1} & \textbf{15.0} & \textbf{16.8} \\
\hline
\end{tabular}%
}
\end{table}

\noindent\textbf{Text-based coordinate injection for effective spatial grounding.}
Cropping the video to the bounding box (\correct~Crop) actually \emph{hurts} accuracy (61.3\%), as it removes the surrounding scene context that the Video LMM relies on for activity and event understanding.
Adding a visual bounding box overlay (\correct~Overlay) recovers accuracy (64.3\%) and improves IoU (13.2\%), but the gains are modest.
In contrast, passing the bounding box coordinates as text (\correct~Coords) achieves the best accuracy--IoU balance (65.4\%, 13.8\%), and combining Overlay with Coords does not improve further (64.5\%, 12.1\%).
This suggests that the Video LMM benefits more from explicit, language-aligned spatial grounding than from visual modifications to the input frames.

\noindent\textbf{More retrieved tracks harm temporal precision.}
Increasing $K$ from 3 to 5 marginally improves average accuracy (65.4\%$\to$66.4\%) but consistently degrades temporal IoU (13.8\%$\to$12.1\%).
We therefore adopt $K{=}3$ as the best accuracy--localization tradeoff.

\noindent\textbf{Sub-task difficulties.}
Across all configurations, \emph{Event} accuracy is consistently the highest (82--88\%), reflecting that event-level questions can often be answered from a single retrieved clip.
\emph{Search} accuracy benefits most from retrieval: \ours{} improves from 49.5\% (VideoLLaMA3) to 60.5\%, confirming that person-centric indexing is the key driver for identity-based queries.
\emph{Activity} is the one category where VideoLLaMA3 remains competitive (61.0\% vs.\ 60.0\%), likely because activity recognition benefits from broader temporal context that retrieval may truncate.
Temporal IoU is uniformly low across all settings (8--12\%), indicating that precise temporal grounding remains an open challenge even with person-centric retrieval.

\noindent\textbf{Analysis of \ours{} variants.}
\ours{}-Global retrieves full video frames instead of person-centric crops to capture more contextual information. This improves overall performance but reduces \emph{search} accuracy in multiple choice (MC) and temporal localization (TL) IoU (57.5\% vs.\ 60.5\% and 14.1\% vs.\ 17.6\%), where identity cues are important. 
This suggests that global indexing benefits scene-level grounding, whereas person retrieval better supports identity-driven queries. 
By combining both person and global retrieval, \ours{}-Hybrid recovers Search performance (59.5\% MC and 16.9\% TL IoU) while leveraging complementary person- and scene-level context.

\begin{table}[t]
\centering
\scriptsize
\small 
\caption{ Comparison across frameworks, retrieval settings, multimodal embeddings, and VLMs.
* For comparable conditions with VideoRAG, which uses 64 frames as input to LLaVA‑Video, we adopt this setting.
MC: multi‑choice; TL: temporal localization.
}
\label{tab:rebuttal}
\setlength{\tabcolsep}{3pt}
\renewcommand{\arraystretch}{1.1}

\resizebox{0.99\textwidth}{!}{
\begin{tabular}{l l l l cc cc cc cc cc}
\toprule
\multirow{3}{*}{Framework} &
\multirow{3}{*}{Setting} &
\multirow{3}{*}{VLM} &
\multirow{3}{*}{Multimodal Encoder} &
\multicolumn{4}{c}{Multimodal (MM)} &
\multicolumn{4}{c}{Text} &
\multicolumn{2}{c}{Overall} \\
\cmidrule(lr){5-8}\cmidrule(lr){9-12}\cmidrule(lr){13-14}
&
&
&
&
\multicolumn{2}{c}{MC(accuracy \%)} &
\multicolumn{2}{c}{TL(IoU)} &
\multicolumn{2}{c}{MC(accuracy \%)} &
\multicolumn{2}{c}{TL(IoU)} &
MC & TL  \\
\cmidrule(lr){5-6}\cmidrule(lr){7-8}\cmidrule(lr){9-10}\cmidrule(lr){11-12}
&
&
&
&
search & avg &
search & avg &
search & avg &
search & avg &
& \\
\midrule
ForeSea &
Person Top3 &
VideoLLaMA3-7B & GCL &
60.5 & 65.4 & 17.6 & 13.8 &
72.0 & 66.7 & 28.5 & 13.3 &
66.0 & 13.6 \\

ForeSea-Global &
Global Top64 &
VideoLLaMA3-7B & GCL &
57.5 & 65.1 & 14.1 & 16.1 &
67.5 & 68.1 & 25.7 & 18.6 &
66.6 & 17.4 \\


ForeSea-Hybrid &
Person Top2 + Global Top24 &
VideoLLaMA3-7B & GCL &
59.5 & 64.6 & 16.9 & 16.8 &
71.0 & 69.1 & 30.5 & 20.1 &
66.9 & \cellcolor{lightgreen}{18.4} \\

\midrule
\multicolumn{14}{c}{Different multimodal encoder experiments} \\
\midrule

ForeSea &
SigLIP &
VideoLLaMA3-7B &
ViT-SO400M-14-SigLIP-384 &
59.0 & 64.5 & 18.4 & 13.2 &
69.0 & 66.6 & 28.9 & 15.8 &
65.5 & 14.5 \\

ForeSea &
SigLIP2 &
VideoLLaMA3-7B &
ViT-SO400M-16-SigLIP2-512 &
59.5 & 64.4 & 17.4 & 14.2 &
70.0 & 67.3 & 29.6 & 15.5 &
65.8 & 14.8 \\

ForeSea &
EVA-CLIP &
VideoLLaMA3-7B &
EVA-CLIP EVA02-E-14-plus &
59.0 & 63.3 & 16.2 & 12.5 &
70.0 & 66.3 & 28.9 & 16.4 &
64.8 & 14.4 \\

\midrule
\multicolumn{14}{c}{Different VLM experiments} \\
\midrule

VideoRAG &
CLIP + APE &
LLaVA-Video-7B-Qwen2 &
CLIP-ViT-large-patch14-336 &
56.5 & 61.9 & 3.1 & 2.8 &
55.5 & 63.8 & 2.2 & 4.3 &
62.9 & \cellcolor{lightred}{3.5} \\

VideoRAG &
CLIP + APE &
VideoLLaMA3-7B &
CLIP-ViT-large-patch14-336 &
73.5 & 67.9 & 3.3 & 9.5 &
72.0 & 68.4 & 11.8 & 10.3 &
68.1 & \cellcolor{lightgreen}{9.9} \\

ForeSea-Hybrid &
Person Top1 + Global Top12* &
LLaVA-Video-7B-Qwen2 &
GCL &
77.0 & 63.3 & 42.0 & 13.5 &
74.0 & 68.6 & 30.8 & 9.3 &
65.9 & \cellcolor{lightred}{11.4} \\

\midrule
\multicolumn{14}{c}{Additional baseline constructed with captions} \\
\midrule

Predicted caption & 
Dense captions by VideoLLaMA3 & 
VideoLLaMA3-7B & 
Caption only & 
58.5 & 52.4 & 1.1 & 8.9 & 
52.0 & 52.2 & 0.4 & 10.2 & 
52.3 & 9.6 \\

Oracle caption & 
Ground-truth UCA dense captions & 
VideoLLaMA3-7B & 
Caption only & 
71.5 & 79.9 & 42.0 & 28.6 & 
92.0 & 86.5 & 28.6 & 33.7 & 
83.2 & 31.1 \\

\bottomrule
\end{tabular}
}
\end{table}

\subsection{Ablation Studies across Encoders and VLMs}
\label{subsec:ablations_vlm}

To assess the inherent difficulty and justification of our benchmark ( if the tasks can be solved without direct visual grounding), we show text-only baselines results.
Also, To evaluate the generalization of our framework, we conduct extensive experiments varying both the multimodal encoders and the foundational Vision-Language Models (VLMs).

\noindent\textbf{Caption-only baselines.}
We evaluate two caption-only baselines by replacing ForeSea's video inputs with timestamped captions. The predicted-caption baseline uses dense captions generated by VideoLLaMA3 from uniformly sampled frames, while the oracle-caption baseline uses the ground-truth UCA captions employed to construct ForeSeaQA as an upper bound baseline.
As shown in Table~\ref{tab:rebuttal}, predicted captions yield the lowest MC accuracy, although their TL performance remains relatively competitive. 
This suggests that generated captions often omit fine-grained visual details, while timestamp information provides useful temporal cues. 
Oracle captions substantially improve overall MC accuracy to 83.2\%, confirming the value of high-quality semantic descriptions. 
However, their TL performance remains limited, indicating that precise temporal grounding is challenging even with oracle captions.

\noindent\textbf{Sensitivity to the multimodal encoder.}
We next investigate whether ForeSea is sensitive to the choice of multimodal retrieval encoder.
In addition to our default GCL encoder, we evaluate SigLIP, SigLIP2, and EVA-CLIP under the same ForeSea retrieval and VideoLLaMA3 inference pipeline.
The four encoders produce similar overall performance.
This show that that the effectiveness of ForeSea does not depend on a particular multimodal embedding model.

\noindent\textbf{Effect of the VLM backbone.}
To disentangle the contribution of the RAG frame work and downstream VLM, we cross-evaluate VideoRAG \cite{luo2024video} and ForeSea using both \colorbox{lightgreen}{\strut VideoLLaMA3} and \colorbox{lightred}{\strut LLaVA‑Video}.
With the same VideoLLaMA3 backbone, ForeSea-Hybrid achieves an overall TL IoU of 18.4\%, compared with 9.9\% for VideoRAG, while maintaining comparable MC accuracy (66.9\% versus 68.1\%).
Thus, ForeSea provides nearly a twofold improvement in temporal localization without relying on a stronger VLM.
The improvement is also preserved when using LLaVA-Video.
Under this backbone, ForeSea-Hybrid improves the overall MC accuracy from 62.9\% to 65.9\% and the TL IoU from 3.5\% to 11.4\% compared with VideoRAG.
Although LLaVA-Video is substantially weaker than VideoLLaMA3 on temporal localization, ForeSea consistently provides a large relative improvement.
These results show that the temporal localization gains primarily arise from the proposed framework rather than from a specific VLM backbone.

\begin{table*}[t]
\centering

\begin{minipage}[t]{0.55\textwidth}
\centering
\footnotesize

\captionof{table}{
\textbf{Inference latency on \oursbench{}.}
TTFT stands for time to first token.
Retrieval, generation, and total time are in seconds;
accuracy and IoU in \%.
}
\label{tab:ablation-latency}

\resizebox{\linewidth}{!}{
\begin{tabular}{lccccc}
\toprule
\multirow{2}{*}{\textbf{Method}}
& \multicolumn{3}{c}{\textbf{Latency (s)}}
& \multicolumn{2}{c}{\textbf{\oursbench{}$^\texttt{MM}$}} \\
\cmidrule(lr){2-4}
\cmidrule(l){5-6}

& \textbf{Retrieval}
& \textbf{Generation$_{\text{(TTFT)}}$}
& \textbf{Total}
& \textbf{Acc}
& \textbf{IoU} \\
\midrule

Qwen2.5-VL-7B-Instruct~\cite{bai2025qwen25vl}
& \textbf{0.0} & 2.1$_{\text{(1.7)}}$ & 2.1 & 58.9 & 8.1 \\

VideoLLaMA3-7B~\cite{zhang2025videollama}
& \textbf{0.0} & 3.8$_{\text{(3.6)}}$ & 3.8 & 61.6 & 10.9 \\

VideoRAG~\cite{luo2024video}$_{\text{ LLaVA-Video-7B-Qwen2}}$
& 2.4 & 2.8$_{\text{(2.3)}}$ & 5.2 & 61.9 & 2.8 \\

T*~\cite{ye2025re}$_{\text{ Qwen2.5-VL-7B-Instruct}}$
& 6.8 & \textbf{0.9}$_{\text{(\textbf{0.6})}}$
& 7.6 & 41.1 & 4.9 \\

\midrule

\ours{}
& \underline{0.5}
& \underline{2.1}$_{\text{(\underline{1.7})}}$
& \underline{2.6}
& \textbf{65.4}
& \underline{13.8} \\

\ours{}-Global
& \underline{0.5}
& \textbf{0.9}$_{\text{(\textbf{0.6})}}$
& \textbf{1.4}
& \underline{65.1}
& \textbf{16.1} \\

\bottomrule
\end{tabular}
}
\end{minipage}
\hfill
\begin{minipage}[t]{0.42\textwidth}
\centering
\footnotesize

\captionof{table}{
\textbf{Performance on open-domain long video benchmarks.}
All numbers are reported from the original papers.
}
\label{tab:LongVideo}

\resizebox{\linewidth}{!}{
\begin{tabular}{lcccc}
\toprule
\textbf{Model}
& \textbf{Param}
& \textbf{Year}
& \textbf{VideoMME}
& \textbf{MLVU} \\
\midrule

LongVU~\cite{shen2024longvu}
& 7B & 2024 & -- & 65.4 \\

LLaVA-Video~\cite{li2024llavaonevision}
& 7B & 2024 & 56.6 & 64.7 \\

TimeMarker~\cite{chen2024timemarker}
& 7B & 2024 & 57.3 & 49.2 \\

InternVL2.5~\cite{chen2024expanding}
& 7B & 2024 & 56.3 & 64.0 \\

Qwen2.5VL~\cite{bai2025qwen25vl}
& 7B & 2025 & 65.1 & 70.2 \\

VideoLLaMA3~\cite{zhang2025videollama}
& 7B & 2025 & 66.2 & 73.0 \\

\midrule

LLaVA-Video + Video-RAG~\cite{luo2024video}
& 7B & 2024 & 58.7 & 72.4 \\

SALOVA-7B~\cite{kim2025salova}
& 7B & 2025 & 53.1 & -- \\

MemVid-7B~\cite{yuan2025memory}
& 7B & 2025 & 63.7 & 58.1 \\

GPT-4o + T*~\cite{ye2025re}
& $>$7B & 2025 & 56.5 & -- \\

LLaVA-OneVision-72B + T*~\cite{ye2025re}
& 72B & 2025 & 59.0 & -- \\

\midrule

\ours{} (Ours)
& 7B & -- & 65.6 & 73.0 \\

\bottomrule
\end{tabular}
}
\end{minipage}

\end{table*}

\subsection{Efficiency Analysis}
\label{subsec:latency}

As shown in Table~\ref{tab:ablation-latency}, \ours{} achieves lower total latency than all baselines while maintaining higher accuracy.
By retrieving only the most relevant person-centric clips, \ours{} reduces the number of frames fed to the Video LMM, directly lowering TTFT and generation time compared to VideoLLaMA3 (which processes the full video).
\ours{} completes inference in 2.6\,s total (1.7\,s TTFT) while achieving the best \oursbench{}$^\texttt{MM}$ accuracy (65.4\%).
In contrast, T$^\ast$ incurs the highest retrieval latency (6.8\,s) despite fast generation, and VideoRAG adds overhead from its dedicated retrieval pipeline (2.4\,s retrieval).

\subsection{Comparison on Existing Benchmarks}
\label{subsec:vidBench_exp}

To assess generalization beyond the surveillance domain, we evaluate \ours{} on three widely used long video benchmarks: VideoMME~\cite{fu2025videomme} MLVU~\cite{zhou2025mlvu}, and LongVideoBench~\cite{wu2024longvideobench}. \footnote{LongVideoBench results are provided in the supplementary material.}
For these benchmarks, \ours{} adapts its database construction: instead of person-centric clips, frames are sampled uniformly at 1\,FPS and indexed at the frame level.
The backbone Video LMM, VideoLLaMA3, supports up to 180 input frames; \ours{} uses at most 90 frames per query (top-60 retrieved + 30 uniformly sampled from the full video).
Despite using only half as many frames, \ours{} achieves comparable performance across all three benchmarks and substantially outperforms prior Video-RAG approaches, as shown in Table~\ref{tab:LongVideo}.


\section{Conclusion}
We introduced \ours{}, a novel Video-RAG framework for forensic search in human surveillance video. \ours{} is, to our knowledge, the first system to handle complex multimodal (image+text) queries and return timestamped, evidence-linked answers, overcoming the limitations of text-only retrieval.
To validate this, we also developed \oursbench{}, the first benchmark for evaluating such temporally-grounded multimodal queries. 
Our experiments demonstrate that \ours{}'s pipeline achieves significant gains in both QA accuracy and temporal IoU over strong baselines. Furthermore, we show our framework's extensibility beyond surveillance, demonstrating its effectiveness on general video understanding tasks. 
This work provides a robust framework and a critical evaluation tool, marking a significant step forward in practical AI forensic analysis.

%
%
\bibliographystyle{splncs04}
\bibliography{main}

\appendix
\section{Introduction}
\label{sec:SuppleIntro}

This supplementary document presents extended experimental results beyond those included in the main paper and additional implementation details on the data generation pipeline. In particular, it contains:

\begin{itemize}
  \item \textbf{Additional experiments with state-of-the-art (SOTA) models:}
    \begin{itemize}
        \item Detailed performance for each sub-task.
        \item Retrieval performance on \oursbench.
        \item Results on LongVideoBench.
    \end{itemize}
    
  \item \textbf{Details of the data generation process:}
    \begin{itemize}
        \item Prompt templates used for dataset construction.
        \item Evaluation metrics and measurement procedures.
    \end{itemize}
\end{itemize}

\section{Additional Experiments}
\label{sec:sup_exp}

\subsection{Analysis of Detailed Subtask Performance}

\begin{table}[h]
\centering
\footnotesize
\caption{Performance comparison on \oursbench{} with subtask details in multimodal (image+text) query}
\label{tab:supp_MM_full}
\resizebox{\textwidth}{!}{%
\setlength{\tabcolsep}{5pt}
\begin{tabular}{l c ccccc ccccc}
\toprule
\multirow{2}{*}{\textbf{Model}} & 
\multirow{2}{*}{\textbf{Params}} &
\multicolumn{5}{c}{\textbf{Multi-choice Accuracy (\%)}} &
\multicolumn{5}{c}{\textbf{Temporal Localization IoU (\%)}} \\
\cmidrule(lr){3-7} \cmidrule(lr){8-12}
& & Search & Activity & Event & Temporal & Avg 
& Search & Activity & Event & Temporal & Avg \\
\midrule
\multicolumn{12}{c}{\textit{Video LMMs (Native)}} \\
\midrule

LLaVA-OneVision~\cite{li2024llavaonevision} 
    & 7B 
    & 54.5 & 54.0 & 76.0 & 40.0 & 56.1
    & 40.8 & 0.5 & 0.1 & 0.1 & 10.4 \\

GLM-4.1V-Thinking~\cite{hong2025glm} 
    & 9B
    & 59.5 & 52.0 & 74.0 & 44.0 & 57.4
    & 38.4 & 0.6 & 0.5 & 0.5 & 10.0 \\

InternVL3~\cite{zhu2025internvl3}
    & 2B
    & 57.0 & 34.0 & 38.0 & 24.0 & 38.3
    & 34.9 & 1.6 & 0.5 & 2.0 & 9.8 \\

InternVL3~\cite{zhu2025internvl3}
    & 8B
    & 63.0 & 54.0 & 83.0 & 45.0 & 61.3
    & 31.1 & 4.1 & 2.6 & 2.8 & 10.2 \\

InternVL3~\cite{zhu2025internvl3}
    & 9B
    & \textbf{64.0} & \textbf{63.0} & 77.0 & 45.0 & 62.3
    & 37.4 & 3.8 & 1.6 & 3.3 & 11.5 \\

VideoLLaMA3~\cite{zhang2025videollama}
    & 7B
    & 49.5 & 61.0 & 83.0 & 53.0 & 61.6
    & 10.7 & 15.0 & 9.8 & 8.2 & 10.9 \\

Qwen2.5-VL~\cite{bai2025qwen25vl}
    & 7B
    & 62.5 & 55.0 & 75.0 & 43.0 & 58.9
    & 25.7 & 3.7 & 0.7 & 2.4 & 8.1 \\

Qwen2.5-VL~\cite{bai2025qwen25vl}
    & 72B
    & \textbf{64.0} & 56.0 & 78.0 & 42.0 & 60.0
    & \textbf{49.2} & 5.3 & 2.2 & 4.4 & 15.3 \\

\midrule
\multicolumn{12}{c}{\textit{Retrieval-Augmented Models (RAG)}} \\
\midrule

VideoRAG~\cite{luo2024video}
    & 7B
    & 56.5 & 58.0 & 85.0 & 48.0 & 61.9
    & 3.1 & 2.0 & 5.4 & 0.8 & 2.8 \\

T$^\ast$~\cite{ye2025re}
    & 7B
    & 52.5 & 30.0 & 50.0 & 32.0 & 41.1
    & 5.4 & 5.4 & 4.7 & 4.2 & 4.9 \\
    
\textbf{\ours{}}
    & 7B
    & 60.5 & 60.0 & 85.0 & 56.0 & \textbf{65.4}
    & 17.6 & 14.1 & 15.4 & 8.2 & 13.8 \\

\textbf{\ours{}-Global}
    & 7B
    & 57.5 & 58.0 & \textbf{88.0} & \textbf{57.0} & 65.1
    & 14.1 & \textbf{18.4} & 19.3 & 12.7 & 16.1 \\

\textbf{\ours{}-Hybrid}
    & 7B
    & 59.5 & 59.0 & 87.0 & 53.0 & 64.6
    & 16.9 & 15.1 & \textbf{20.1} & \textbf{15.0} & \textbf{16.8} \\
    
\bottomrule
\end{tabular}
}
\end{table}

\begin{table*}[h]
\centering
\footnotesize
\caption{
Performance comparison of state-of-the-art Video LMMs and RAG models on \oursbench{} using text queries.}
\label{tab:supp_text_full}
\resizebox{\textwidth}{!}{
\setlength{\tabcolsep}{5pt}
\begin{tabular}{l c ccccccc ccccccc}
\toprule
\multirow{2}{*}{\textbf{Model}} &
\multirow{2}{*}{\textbf{Params}} &
\multicolumn{7}{c}{\textbf{Multi-choice Accuracy (\%)}} &
\multicolumn{7}{c}{\textbf{Temporal Localization IoU (\%)}} \\
\cmidrule(lr){3-9} \cmidrule(lr){10-16}
& & Search & Activity & Event & Temporal & Counting & Anomaly & Avg
& Search & Activity & Event & Temporal & Counting & Anomaly & Avg \\
\midrule
\multicolumn{16}{c}{\textit{Video LMMs (Native Models)}} \\
\midrule

LLaVA-OneVision~\cite{li2024llavaonevision}
& 7B
& 60.0 & 54.0 & 76.0 & 39.0 & 45.9 & 75.9 & 58.5
& \textbf{44.1} & 1.1 & 0.0 & 0.5 & 0.0 & 0.3 & 7.7 \\

GLM-4.1V-Thinking~\cite{hong2025glm}
& 9B
& 64.5 & 48.0 & 77.0 & 43.0 & 35.1 & 63.0 & 55.1
& 43.9 & 3.7 & 0.6 & 0.4 & 1.3 & 0.4 & 8.4 \\

InternVL3~\cite{zhu2025internvl3}
& 2B
& 60.0 & 34.0 & 31.0 & 25.0 & 27.0 & 27.8 & 34.1
& 35.7 & 2.4 & 0.3 & 1.5 & 2.5 & 0.4 & 7.1 \\

InternVL3~\cite{zhu2025internvl3}
& 8B
& 67.0 & 49.0 & 90.0 & 46.0 & 43.2 & \textbf{88.9} & 63.3
& 41.1 & 5.1 & 4.0 & 4.5 & 2.9 & 1.6 & 9.9 \\

InternVL3~\cite{zhu2025internvl3}
& 9B
& 66.0 & \textbf{60.0} & 81.0 & 51.0 & 35.1 & 83.0 & 62.7
& 39.1 & 6.3 & 1.3 & 3.5 & 1.6 & 0.9 & 8.8 \\

VideoLLaMA3~\cite{zhang2025videollama}
& 7B
& 63.5 & 59.0 & 90.0 & 57.0 & \textbf{56.8} & 79.6 & 67.7
& 29.4 & 18.1 & 11.1 & 12.9 & 11.7 & 9.7 & 15.5 \\

VideoLLaMA3~\cite{zhang2025videollama}
& 2B
& 47.5 & 55.0 & 89.0 & 44.0 & 45.9 & 87.0 & 61.4
& 27.6 & 5.3 & 1.8 & 9.1 & 0.0 & 0.2 & 7.3 \\

Qwen2.5-VL~\cite{bai2025qwen25vl}
& 7B
& 70.5 & 51.0 & 82.0 & 42.0 & 32.4 & 75.9 & 59.0
& 36.1 & 4.2 & 0.8 & 2.3 & 1.4 & 0.0 & 7.5 \\

Qwen2.5-VL~\cite{bai2025qwen25vl}
& 72B
& 66.0 & 48.0 & 81.0 & 44.0 & 45.9 & 83.3 & 61.4
& 38.0 & 6.3 & 2.3 & 4.7 & 6.7 & 2.8 & 10.1 \\

\midrule
\multicolumn{16}{c}{\textit{Retrieval-Augmented Models}} \\
\midrule

VideoRAG~\cite{luo2024video}
& 7B
& 55.5 & 59.0 & 91.0 & 51.0 & 43.2 & 83.3 & 63.8
& 2.2 & 2.6 & 10.3 & 1.9 & 5.9 & 2.6 & 4.3 \\

T$^\ast$~\cite{ye2025re}
& 7B
& 61.0 & 37.0 & 70.0 & 40.0 & 27.0 & 55.6 & 48.4
& 4.9 & 6.6 & 4.1 & 4.3 & 1.9 & 3.3 & 4.2 \\

\textbf{\ours{}}
& 7B
& \textbf{72.0} & 56.0 & 91.0 & \textbf{62.0} & 43.2 & 75.9 & 66.7
& 28.5 & 11.4 & 16.0 & 9.4 & 7.2 & 7.3 & 13.3 \\

\textbf{\ours{}-Global}
& 7B
& 67.5 & 56.0 & \textbf{93.0} & 59.0 & 51.4 & 81.5 & 68.1
& 25.7 & \textbf{19.0} & 18.5 & 14.0 & 20.1 & \textbf{14.4} & 18.6 \\

\textbf{\ours{}-Hybrid}
& 7B
& 71.0 & 55.0 & \textbf{93.0} & 61.0 & 51.4 & 83.3 & \textbf{69.1}
& 30.5 & 17.1 & \textbf{21.7} & \textbf{16.2} & \textbf{23.9} & 11.1 & \textbf{20.1} \\

\bottomrule
\end{tabular}
}
\end{table*}

To further analyze our framework, we present detailed sub-task performance across multimodal and text-only queries in Tables~\ref{tab:supp_MM_full} and \ref{tab:supp_text_full}, respectively, comparing \ours{} with state-of-the-art Video LMMs and RAG models. 
Both tables extend the results in Table~1 of the main paper.

Table~\ref{tab:supp_MM_full} focuses on the highly challenging multimodal setting that closely reflects real-world forensic search scenarios. 
\ours{} achieves the highest overall multi-choice accuracy (65.4\%), outperforming 72B-parameter general-purpose Video LMMs as well as all RAG baselines. 
By centering retrieval on human subjects, \ours{} effectively suppresses background noise and excels in complex reasoning tasks such as Activity (60.0\%) and Event (85.0\%) recognition. Meanwhile, \ours{}-Global achieves stronger temporal localization performance, reaching 16.1\% average IoU. 
Combining the strengths of both approaches, \ours{}-Hybrid attains the best overall balance between reasoning and localization, achieving state-of-the-art performance across most metrics and the highest average scores among all evaluated methods.

In Table~\ref{tab:supp_text_full}, all three variants substantially outperform existing RAG approaches (e.g., VideoRAG and T$^\ast$). \ours{} achieves the highest Search accuracy (72.0\%) and attains 13.3\% average IoU, more than tripling the performance of VideoRAG (4.3\%). \ours{}-Global further improves temporal grounding, achieving 18.6\% average IoU and strong overall accuracy (68.1\%). Notably, \ours{}-Hybrid delivers the best overall performance, achieving state-of-the-art results with 69.1\% average multi-choice accuracy and 20.1\% average IoU. 
It attains the highest localization performance on Event (21.7\%), Temporal (16.2\%), and Counting (23.9\%) tasks while maintaining competitive reasoning accuracy across categories.

Overall, the proposed variants consistently outperform VideoRAG-based methods and general Video LMMs. While the streamlined architecture of \ours{}-Global provides strong holistic video understanding and temporal grounding, \ours{} offers more precise reasoning for search-centric queries. 
By integrating both local person-centric retrieval and global contextual evidence, \ours{}-Hybrid achieves the most favorable trade-off, delivering the strongest overall performance across both reasoning and temporal localization tasks.

\subsection{Comparing Multimodal Embeddings for Video Retrieval}

\begin{table}[h]
\centering
\resizebox{\textwidth}{!}{
\setlength{\tabcolsep}{8pt}
\renewcommand{\arraystretch}{1.25}
\begin{tabular}{lcccccccccccc}
\toprule
& \multicolumn{3}{c}{\textbf{Top1}}
& \multicolumn{3}{c}{\textbf{Top3}}
& \multicolumn{3}{c}{\textbf{Top5}}
& \multicolumn{3}{c}{\textbf{Top10}} \\
\cmidrule(lr){2-4} \cmidrule(lr){5-7} \cmidrule(lr){8-10} \cmidrule(lr){11-13}
& @0 & @0.1 & @0.3
& @0 & @0.1 & @0.3
& @0 & @0.1 & @0.3
& @0 & @0.1 & @0.3 \\
\midrule

\rowcolor{gray!10}
\multicolumn{13}{l}{\textbf{Query Text}} \\

CLIP
& 47.9 & 29.12 & 11.49
& 72.3 & 49.43 & 24.33
& 80.7 & 57.47 & 30.65
& 87.37 & \textbf{65.90} & \textbf{38.89} \\

SigLIP ViT-SO400M-14-SigLIP-384
& 41.1 & 24.90 & 13.03
& 75.6 & 49.23 & 26.05
& 85.5 & 58.62 & \textbf{32.95}
& 90.22 & 65.71 & 37.36 \\

ViT-SO400M-16-SigLIP2-512
& 47.5 & 29.69 & 13.22
& 75.4 & 48.28 & 25.67
& 84.9 & 58.05 & 32.76
& 90.22 & 64.75 & 36.97 \\

EVA-CLIP EVA02-E-14-plus
& \textbf{56.4} & \textbf{35.25} & \textbf{17.05}
& \textbf{81.7} & \textbf{54.02} & \textbf{28.35}
& \textbf{88.8} & \textbf{60.92} & 32.76
& \textbf{91.04} & 65.71 & 36.21 \\

GCL (ours)
& 52.1 & 34.48 & 13.98
& 73.9 & 50.38 & 23.56
& 82.7 & 57.66 & 31.23
& 87.17 & 65.33 & 37.74 \\

\midrule

\rowcolor{gray!10}
\multicolumn{13}{l}{\textbf{Query Multimodal}} \\

CLIP
& 41.4 & 30.6 & 9.2
& 69.7 & 52.0 & 21.9
& 75.5 & 58.3 & 28.5
& 85.2 & 68.1 & 40.6 \\

SigLIP ViT-SO400M-14-SigLIP-384
& 57.5 & 38.02 & 20.51
& 84.0 & 56.22 & 29.49
& \textbf{90.0} & 62.90 & 35.25
& \textbf{93.56} & 68.89 & 40.78 \\

ViT-SO400M-16-SigLIP2-512
& 60.9 & \textbf{42.40} & 21.43
& 84.0 & 56.91 & \textbf{30.18}
& 89.5 & 62.90 & 35.94
& \textbf{93.56} & \textbf{69.12} & 40.09 \\

EVA-CLIP EVA02-E-14-plus
& \textbf{66.4} & 41.24 & \textbf{22.12}
& \textbf{86.9} & 57.37 & 29.49
& \textbf{90.0} & \textbf{63.13} & \textbf{36.41}
& 93.32 & 68.66 & 39.63 \\

GCL (ours)
& 55.4 & 37.7 & 12.7
& 76.8 & \textbf{58.3} & 26.9
& 81.8 & 63.1 & 34.0
& 87.1 & 69.1 & \textbf{41.9} \\
\bottomrule
\end{tabular}
}
\end{table}



We further analyze the retrieval component of \ours{} by comparing GCL \cite{gcl} and several representative vision-language retrieval models, including CLIP \cite{clip}, SigLIP \cite{zhai2023sigmoid}, SigLIP2 \cite{tschannen2025siglip2}, and EVA-CLIP \cite{sun2023evaclip}, on \oursbench{} under both multimodal and text-only query settings. 
We adopt GCL as our retrieval backbone by following the framework described in Section 4.2 of the main paper, embedding human‑centric video clips. 
Because \ours{} depends on retrieval to narrow down the candidate clips before VideoLMM-based reasoning, retrieval quality is crucial to overall system performance.

Across most metrics and in both query modalities, modern vision-language encoders substantially outperform the original CLIP baseline. In particular, EVA-CLIP achieves the strongest overall retrieval performance, while SigLIP and SigLIP2 consistently improve retrieval accuracy under both multimodal and text-only queries. 
The performance gains are especially pronounced for multimodal queries and small‑K retrieval (Top‑1 and Top‑3), suggesting that stronger image-text alignment directly benefits human-centric video retrieval.

As shown in Table~3 of the main paper, ForeSea with SigLIP, SigLIP2, and EVA-CLIP generally achieves higher retrieval accuracy than the GCL-based setting. 
These results indicate that retrieval performance remains a key bottleneck of \ours{} and suggest that replacing the current retrieval backbone with more powerful multimodal embedding models could further improve the end-to-end performance of the ForeSea pipeline.

\subsection{Evaluating ForeSea on LongVideoBench}

\begin{table}[h]
\centering
\caption{LongVideoBench results}
\label{tab:LongVid}
\footnotesize

\resizebox{0.55\linewidth}{!}{
\begin{tabular}{lccc}
\hline
\textbf{Model} & \textbf{Param} & \textbf{Year} & \textbf{LongVid} \\
\hline
LongVU \cite{shen2024longvu}                   & 7B & 2024 & 59.5 \\
LLaVA-Video \cite{li2024llavaonevision}        & 7B & 2024 & 58.2 \\
TimeMarker \cite{chen2024timemarker}           & 7B & 2024 & 56.3 \\
InternVL2.5 \cite{chen2024expanding}           & 7B & 2024 & 54.6 \\
Qwen2.5VL \cite{bai2025qwen25vl}               & 7B & 2025 & 54.7 \\
VideoLLaMA3 \cite{zhang2025videollama}         & 7B & 2025 & 59.8 \\
\hline
Video-RAG (7B) \cite{luo2024video} & 7B & 2024     & 45.0 \\
SALOVA-7B \cite{kim2025salova}                 & 7B & 2025     & 44.6 \\
MemVid-7B \cite{yuan2025memory}                & 7B & 2025     & 44.4 \\
\hline
\ours{} (Ours) wo subtitle                                 & 7B & --       & 63.5 \\
\ours{} (Ours) with subtitle                                 & 7B & --       & 65.0 \\
\hline
\end{tabular}}
\end{table}

To evaluate the generalization ability of \ours{} beyond surveillance videos, we report results on LongVideoBench using the same retrieval setting as in Table~4, which is top-60 retrieved frames together with 30 uniformly sampled frames. 
\ours{} achieves the strongest LongVideoBench score among the compared 7B models. 
In particular, it outperforms recent VideoLMM baselines such as LongVU, LLaVA-Video, TimeMarker, InternVL2.5, Qwen2.5VL, and VideoLLaMA3, and also exceeds prior retrieval-based methods including Video-RAG, SALOVA, and MemVid. 
This is a meaningful result because it shows that the benefit of \ours{} is not restricted to the surveillance domain.

The strong transfer performance suggests that \ours{}'s main advantage comes from its ability to identify compact and relevant evidence before passing it to the VideoLMM. Rather than relying on dense processing of the full video, the method focuses the generator on a smaller set of informative content, which improves both scalability and reasoning quality. 
Therefore, the LongVideoBench result provides additional evidence that \ours{} is a generally useful framework for long-video understanding, not only a benchmark-specific solution for \ours{}QA.

\newtcolorbox{promptbox}[2][]{%
  colback=gray!5!white,      
  colframe=gray!75!black,    
  coltitle=white,            
  title=\textbf{#2},         
  fonttitle=\bfseries\sffamily,
  breakable,                 
  enhanced,
  attach boxed title to top left={yshift=-2mm, xshift=2mm},
  boxrule=0.5pt,
  arc=2mm,
  left=2mm, right=2mm, top=2mm, bottom=2mm,
  #1
}

\section{Details of \oursbench{} benchmark}
\label{sec:sup_data}

\subsection{Task Formulation}
We design the following 6 subtasks that incorporate temporal grounding and multimodal queries in a multiple-choice format, with different levels of reasoning required in the LMM:
\begin{itemize}
    \item \textbf{Search} (\textbf{SE}): Needle-in-a-haystack questions that require the model to accurately localize a queried person of interest in time.
    To ensure a balanced dataset, we match each positive query with a \emph{negative} one by pairing the same question with a video where the target (person or moment) is absent. 
    
    \item \textbf{Event} (\textbf{EV}): Questions about events involving multiple individuals in the scene, requiring the model to understand group activities and human-to-human interactions.   
    
    \item \textbf{Activity} (\textbf{AC}): Questions about activities of specific individuals that require the model to perform action recognition and retrieval in the surveillance video. 
    
    \item \textbf{Temporal} (\textbf{TM}): Questions about multiple activities or sequences of events. This tests the model's ability to understand and reason about temporal relationships and broader context across multiple moments.
    
    \item \textbf{Counting} (\textbf{CT}): Questions that ask for the number of people or events in the video.
    This requires the model to aggregate and recall all instances relevant to the query in order to answer correctly.
    
    \item \textbf{Anomaly} (\textbf{AN}): Questions about abnormal or unusual events in the video.
    This requires a holistic understanding of the situation to detect and locate moments of anomaly.
\end{itemize}

\subsection{Data Generation Prompts}
\label{subsec:prompts}

To ensure reproducibility and transparency, we provide the exact prompt templates used to generate our dataset. We employ a Large Language Model (LLM) to process dense video captions and synthesize high-quality Question-Answer (QA) pairs. 

To achieve diversity in the dataset, we designed specific prompts for six distinct task categories: \textbf{Activity Understanding, Anomaly Detection, Counting, Group Events, Person Search, and Temporal Reasoning}. Each prompt includes a system instruction, strict JSON input/output definitions, and few-shot examples to guide the generation process. The specific templates are detailed below.

\begin{promptbox}{Prompt Template 1: Activity Understanding}
\textbf{1. Activity}

You are given a list of dense video captions with timestamps and a single person reference extracted from those captions. The person reference is a description of a person based on their appearance and/or activity. Your task is to generate up to 3 multiple-choice, activity-focused QA pairs that help identify or verify what this person did in the video.

\textit{Note: The captions are used to generate questions/distractors. The correct answer is derived from the video content.}

\textbf{Input Format:}
You will receive a JSON object:
\begin{lstlisting}[style=jsonstyle]
{
  "captions": [
    { "start": float, "end": float, "text": string }, ...
  ],
  "person_reference": string
}
\end{lstlisting}

\textbf{Output Format:}
Return a JSON array. Each entry must include:
\begin{itemize}
    \item "question": Identifying/verifying activity.
    \item "answer": Concise answer derived from video.
    \item "distractors": 3 plausible but incorrect alternatives.
    \item "person": The "person\_reference" string.
    \item "timestamp": \{ "start": float, "end": float \}.
\end{itemize}

\textbf{Guidelines:}
\begin{itemize}
    \item Focus only on the person described in "person\_reference".
    \item Use caption text to infer activity-based questions.
    \item Distractors should be plausible (e.g., actions by others).
    \item Return empty list if no activity info is available.
\end{itemize}

\textbf{Input Template:}
\begin{lstlisting}[style=jsonstyle]
{{ input_dict | tojson }}
\end{lstlisting}
\end{promptbox}

\begin{promptbox}{Prompt Template 2: Anomaly Detection}
\textbf{2. Anomaly}

You are given a list of dense video captions with timestamps. Your task is to generate up to 3 multiple-choice QA pairs that require \textbf{global understanding} and focus specifically on \textbf{anomaly detection}.

These questions should focus on: identifying unusual events, locating abnormal behaviors, or recognizing inconsistencies.

\textbf{Output Format:}
Return a JSON array where each entry includes:
\begin{itemize}
    \item "question": Anomaly detection question.
    \item "answer": Description of the unusual event.
    \item "distractors": 3 plausible events that did not occur.
    \item "timestamp": \{ "start": float, "end": float \}.
\end{itemize}

\textbf{Examples of Questions:}
\begin{itemize}
    \item "Which of the following describes the unusual event?"
    \item "What unexpected behavior was observed?"
\end{itemize}

\textbf{Input Template:}
\begin{lstlisting}[style=jsonstyle]
{{ input_dict | tojson }}
\end{lstlisting}
\end{promptbox}

\begin{promptbox}{Prompt Template 3: Counting}
\textbf{3. Counting}

Your task is to generate up to 3 multiple-choice QA pairs that require \textbf{global understanding} and focus specifically on \textbf{counting-type questions} (e.g., event frequency, object count).

\textbf{Output Format:}
Return a JSON array where each entry includes:
\begin{itemize}
    \item "question": A counting question.
    \item "answer": Correct answer string (must include count).
    \item "distractors": 3 incorrect counts.
    \item "timestamps": A \textbf{list} of timestamp objects for each instance.
\end{itemize}

\textbf{Guidelines:}
\begin{itemize}
    \item Distractors should be plausible numbers (e.g., close to the real count).
    \item Timestamps should correspond to each instance that contributes to the count.
\end{itemize}

\textbf{Input Template:}
\begin{lstlisting}[style=jsonstyle]
{{ input_dict | tojson }}
\end{lstlisting}
\end{promptbox}

\begin{promptbox}{Prompt Template 4: Event Understanding}
\textbf{4. Event}

You are given captions and a person reference. Your task is to generate up to 3 QA pairs focusing on \textbf{group events} (e.g., sports, protests, fights). Questions should help understand:
1. The \textbf{role} of the person (participant, instigator, etc.).
2. The \textbf{development} of the event.

\textbf{Output Format:}
JSON array containing "question", "answer", "distractors", "person", and "timestamp".

\textbf{Question Examples:}
\begin{itemize}
    \item "What role did <person> play in <event>?"
    \item "How did <person> contribute to the escalation?"
\end{itemize}

\textbf{Input Template:}
\begin{lstlisting}[style=jsonstyle]
{{ input_dict | tojson }}
\end{lstlisting}
\end{promptbox}

\begin{promptbox}{Prompt Template 5: Person Search}
\textbf{5. Search}

Your task is to generate insightful QA pairs that focus specifically on searching for a person in the video based on their description.

\textbf{Output Format:}
Return a JSON array where each entry includes:
\begin{itemize}
    \item "question": Refers to person using full description.
    \item "question\_indirect": Refers to person \textbf{without} mentioning appearance (e.g., "this person").
    \item "answer": Accurate answer derived from video.
    \item "person": The person reference string.
    \item "timestamp": \{ "start": float, "end": float \}.
\end{itemize}

\textbf{Rules:}
\begin{itemize}
    \item If reference is appearance-based: Indirect question must not mention clothing/hair.
    \item If reference is activity-based: Indirect question may mention activity.
\end{itemize}

\textbf{Input Template:}
\begin{lstlisting}[style=jsonstyle]
{{ input_dict | tojson }}
\end{lstlisting}
\end{promptbox}

\begin{promptbox}{Prompt Template 6: Temporal Reasoning}
\textbf{6. Temporal}

Your task is to generate up to 3 QA pairs focusing on the \textbf{sequence of events}. Questions should address:
1. What happened \textbf{before or after} an event.
2. The \textbf{temporal relationship} between actions.
3. The \textbf{order} of activities.

\textbf{Output Format:}
Standard JSON array with question, answer, distractors, person, and timestamp.

\textbf{Question Examples:}
\begin{itemize}
    \item "What did <person> do before <event>?"
    \item "Which of the following activities did <person> do first?"
\end{itemize}

\textbf{Input Template:}
\begin{lstlisting}[style=jsonstyle]
{{ input_dict | tojson }}
\end{lstlisting}
\end{promptbox}

\subsection{Surveillance-specific analysis}
\label{sec:supp_analysis}

We conduct three analyses on ForeSeaQA\textsubscript{MM}.
\textbf{(a) Resolution:} All videos are low-resolution (320$\times$240), reflecting realistic surveillance conditions across the entire benchmark.
\textbf{(b) Query image size:} We group samples by the relative size of the query image within the frame. 
Performance remains stable across ( small / medium / large ) targets (Acc: 64.7 / 65.5 / 62.4 for crops occupying 2–9\% / 9–17\% /  17–50\% of the frame), demonstrating robustness to scale variation.
\textbf{(c) Scene type:} Videos span diverse scenes, including indoor (69\%, Acc=64.3, IoU=13.9), outdoor (24\%, Acc=60.0, IoU=16.1), and parking/garage (7\%, Acc=77.1, IoU=16.1).
We also note that ForeSea targets underexplored yet realistic scenarios, particularly multimodal question answering and temporal grounding.

\subsection{Evaluation Metrics}
\label{sec:supp_metrics}

We evaluate both the \textbf{retrieval} and \textbf{\oursbench{}} tasks using metrics designed to assess semantic correctness as well as temporal grounding quality.

\paragraph{Retrieval Metrics.}
Each retrieval query is associated with a ground-truth temporal interval. A retrieved segment is considered correct if its predicted temporal span sufficiently overlaps with the ground-truth event.

\textbf{Top-$K$@IoU.}
To assess temporal precision, we report Top-$K$@IoU, which measures whether any of the top-$K$ retrieved segments achieves an intersection-over-union (IoU) with the ground-truth interval exceeding a threshold~$\tau$. For a retrieved interval $R$ and ground-truth interval $G$, the temporal IoU is defined as
\[
\mathrm{IoU}(R, G) = \frac{|R \cap G|}{|R \cup G|}.
\]
We report results for $\tau \in \{0, 0.1, 0.3\}$. Top-$K$@0 indicates whether the retrieved interval overlaps the ground-truth event in any way, while Top-$K$@0.1 and Top-$K$@0.3 require increasingly stringent temporal alignment.

\paragraph{\oursbench{} Metrics.}
The \oursbench{} benchmark includes both \textbf{binary} (yes/no) and \textbf{multiple-choice} questions. Binary questions appear only in the \textit{search} subtask; all other subtasks use a multiple-choice format.

\textbf{Accuracy.}
We use classification accuracy as the primary evaluation metric, defined as the percentage of questions for which the model predicts the correct answer. This metric is used across all QA subtasks.

\textbf{Temporal IoU.}
In addition to answer accuracy, we evaluate whether the predicted temporal evidence aligns with the ground-truth time range. For a predicted interval $\hat{G}$ and ground-truth interval $G$, temporal IoU is computed as above. For the binary search task, where the model may predict that no relevant event is present, we adopt the following conventions:
\begin{enumerate}
    \item If the ground truth is negative but the model predicts a positive event, the temporal IoU is set to $0$.
    \item If both the ground truth and the prediction are negative, the temporal IoU is set to $1$.
\end{enumerate}

Overall, these metrics provide complementary perspectives: retrieval metrics evaluate whether the relevant evidence is successfully retrieved and temporally grounded, while QA metrics measure both answer correctness and the quality of temporal localization.

\end{document}